\documentclass[11pt]{article}

\usepackage[preprint]{acl}
\usepackage{times}
\usepackage{latexsym}

\usepackage[T1]{fontenc}

\usepackage[utf8]{inputenc}

\usepackage{microtype}

\usepackage{inconsolata}

\usepackage{graphicx}
\usepackage{xcolor}

\usepackage{multirow, makecell}

\usepackage{amsmath}
\usepackage{amssymb}

\usepackage{subcaption}

\usepackage{enumitem}

\usepackage{pifont}
\usepackage{cprotect}

\usepackage{listings}
\lstnewenvironment{code}[1][]%
  {\noindent\minipage{\linewidth}\medskip 
   \lstset{basicstyle=\ttfamily,frame=single,#1}}
  {\endminipage}

\definecolor{Green}{rgb}{0.0, 0.5, 0.0}
\definecolor{red-brown}{rgb}{0.8, 0.1, 0.1}

\newcommand{\incorrect}[1]{\textcolor{red-brown}{#1}}

\title{Improving Visual Sensitivity of LLMs on Multimodal Machine Translation with Metric-based Loss Weighting}

\author{
 \textbf{Paweł Mąka}\textsuperscript{1},
 \textbf{Piotr Andruszkiewicz}\textsuperscript{2, 3},
 \textbf{Yusuf Can Semerci}\textsuperscript{1},
 \textbf{Jan Scholtes}\textsuperscript{1}, \\
 \textbf{Gerasimos Spanakis}\textsuperscript{1}
\\
\\
 \textsuperscript{1}Department of Advanced Computing Sciences, Maastricht University\\
 \textsuperscript{2}Warsaw University of Technology\\
 \textsuperscript{3}IDEAS Research Institute
\\
    \texttt{\{pawel.maka, y.semerci, j.scholtes, jerry.spanakis\}@maastrichtuniversity.nl} \\
    \texttt{piotr.andruszkiewicz@pw.edu.pl}
}

\begin{document}
\maketitle
\begin{abstract}
Multimodal Machine Translation aims to incorporate additional signal from non-textual modalities to improve translations by resolving ambiguities.
While models, through multimodal fusion, are able to accept images related to the source text, they can ignore this information. Therefore, increasing their visual sensitivity remains an active research area.
In this work, we introduce a training method, Metric-based Loss Weighting, that improves visual grounding of translations by increasing the loss function for tokens that benefit from the accompanying image. We identify these tokens using the Point-wise Cross-mutual Information (PCXMI) metric, which compares the model's output probabilities with and without visual context. We introduce a Congruency-based PCXMI metric and experimentally show that both metrics working in combination yield the best results.
We evaluate our method by fine-tuning three pretrained Multimodal Large Language Models on the task of Image-guided Machine Translation for three language directions. 
Metric-based Loss Weighting outperforms other tested methods on the CoMMuTE contrastive dataset, improving accuracy by up to more than 7 percentage points compared to standard fine-tuning, while maintaining strong general translation performance.

\end{abstract}

\section{Introduction}

Sentence-level Machine Translation (MT) systems achieved high performance~\citep{vaswani2017attention, hassan2018achieving, costa2022no, tiedemann2024democratizing}, but they are inherently limited by the lack of context. In Multimodal Machine Translation (MMT), the additional signal from other modalities can improve translation in cases where text alone is ambiguous. This signal can be audio, video, or image~\citep{sulubacak2020multimodal, garg-etal-2022-multimodality, feng-etal-2025-multimodal}. In this study, we concentrate our efforts on Image-based MT.

\begin{figure*}
    \centering
    \includegraphics[width=0.9\linewidth]{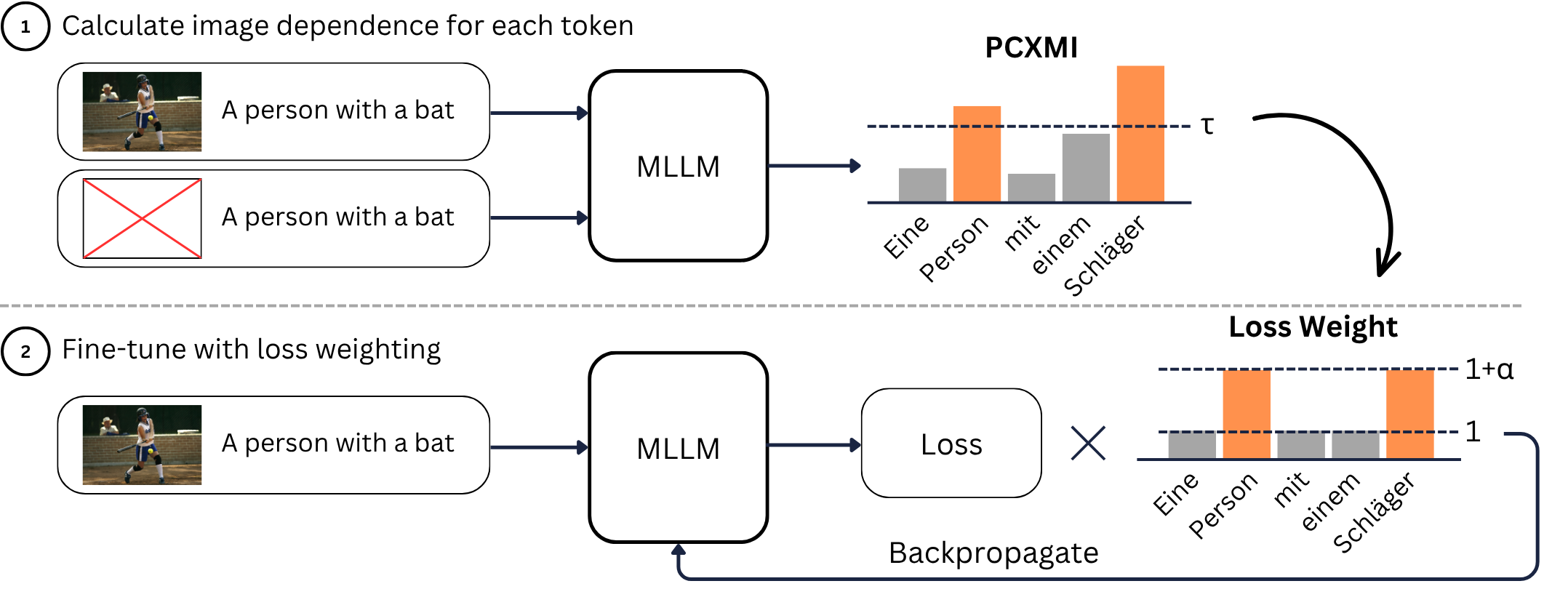}
    \caption{The illustration of Metric-based Loss Weighting: (1) Calculate the image dependence metric for each token (PCXMI), based on the model's output probabilities with and without an image (or with an incongruent image); (2) Fine-tune the model with increased loss (by $\alpha$) for the identified tokens (with PCXMI larger than $\tau$).}
    \label{fig:method}
\end{figure*}

While the images are available to the MMT systems, the reliance on the other modality is not guaranteed. Indeed, \citet{elliott-2018-adversarial} did not find the expected performance drop of the tested systems after randomly replacing images. \citet{wu-etal-2021-good} attributed some gains of the MMT systems to regularization.
Therefore, visual sensitivity of the MMT models has seen increased attention from researchers in recent years~\citep{sato-etal-2023-choosing, zhu-etal-2023-beyond, cheng-etal-2024-soul, bowen-etal-2024-detecting, futeral-etal-2025-towards}. 

As traditional translation metrics such as BLEU~\citep{papineni-etal-2002-bleu} are not sensitive enough to judge disambiguations, we turn towards CoMMuTE~\citep{futeral-etal-2023-tackling}. It is a contrastive dataset where the tested models are presented with multiple possible translations of the ambiguous source sentence and tasked with selecting the correct one. The selection is based on the probability that models give to each option, and the final metric is the accuracy of assigning the highest probability to the correct one.

The field of MMT went from training specialized architectures from random initialization~\citep{yin-etal-2020-novel, liu2021gumbel} to using pre-trained models as building blocks and fine-tuning the resulting model with multimodal parallel data~\citep{gupta2023cliptrans, yu-etal-2025-imagination}. Recently, more focus has been given to Multimodal Large Language Models (MLLMs)~\citep{alayrac2022flamingo, liu2023visual, laurencon2023obelics, Qwen2.5-VL} due to their generalizability and strong performance in different tasks~\citep{zhu-etal-2024-multilingual}. In this study, we concentrate on methods that could be used to enhance fine-tuning of MLLMs on the task of MMT to increase the visual grounding of the translation.

We argue that the insufficient number of examples in the widely used MMT training datasets that require images for correct translation is causing the training signal to be sparse and, therefore, not enough for the models to learn to rely on context~\citep{futeral-etal-2023-tackling}. To mitigate this issue, we propose a training strategy where we identify the tokens in the target sequence that benefit from context and, subsequently, increase the loss for those tokens during training. To identify the image-dependent tokens, we use the Point-wise Cross-Mutual Information (PCXMI) metric~\citep{fernandes-etal-2023-translation}, which compares the output probabilities of the model in two situations: with and without the context (see Figure~\ref{fig:method}). We propose a modification to this metric by using an incongruent image as context instead of the lack of it. We combine both metrics to achieve the best results. We test our method against several techniques adapted to fine-tuning of MLLMs using three models and three language directions. Our Metric-based Loss Weighting method outperforms the baselines in terms of CoMMuTE accuracy on most of the tested scenarios. It improves the accuracy by up to more than 7 percentage points compared to the standard fine-tuning.

Our contributions are as follows:
\begin{enumerate}[topsep=3pt,itemsep=3pt,partopsep=0pt, parsep=0pt, itemindent=15pt, leftmargin=0pt]
    \item We introduce a training strategy for fine-tuning MLLMs that improves their visual sensitivity on the task of MMT based on identifying and increasing loss function for image-dependent tokens. 
    \item We adapt the Point-wise Cross-Mutual Information (PCXMI) metric~\citep{fernandes-etal-2023-translation} to MMT, introduce Congruency-based PCXMI metric, and experimentally show that both metrics working in combination yield best results.
    \item We adapt and evaluate several existing methods for fine-tuning MLLMs on the task of MMT, establishing the baseline for future research.
    \item We show that the pre-trained models exhibit high visual sensitivity which can be further increased with specialized fine-tuning techniques.
    \item We empirically confirm the discrepancy between traditional translation metrics such as BLEU~\citep{papineni-etal-2002-bleu} and the image reliance of the MLLMs.
\end{enumerate}

\section{Related Work}


Researchers have been interested in incorporating the information coming from other modalities to enhance text-to-text translation~\citep{sulubacak2020multimodal, garg-etal-2022-multimodality, feng-etal-2025-multimodal}. While this includes audio and video~\citep{gu-etal-2021-video, li-etal-2023-video}, large focus has been concentrated on single images related to the source text~\citep{elliott-etal-2017-findings, barrault-etal-2018-findings, cheng-etal-2024-soul, futeral-etal-2025-towards, yu-etal-2025-imagination}. The methods proposed specialized architectures concentrating on feature extraction~\citep{yao-wan-2020-multimodal, fang-feng-2022-neural}, cross-modal fusion and alignment~\citep{9664333, guo-etal-2022-lvp}, and cross-modal attention~\citep{nishihara-etal-2020-supervised, gupta2023cliptrans}.

Other notable directions include: employing auxiliary tasks, such as Visually Conditioned Masked Language Modeling (VMLM)~\citep{futeral-etal-2023-tackling, futeral-etal-2025-towards}, image captioning~\citep{gupta2023cliptrans}, or visual question answering~\citep{zhang-etal-2025-vqa, gao-etal-2025-multimodal}, masking the source sentence to increase visual grounding~\citep{sato-etal-2023-choosing, cheng-etal-2024-soul, bowen-etal-2024-detecting}, and data augmentation by image generation~\citep{long-etal-2021-generative, li2022valhalla, chen-etal-2025-make}, using text-to-image models~\cite{goodfellow2014generative, 9878449}, or backtranslation~\cite{li-etal-2021-vision}.

The MMT models are often trained on the widely used Multi30k dataset and evaluated on the accompanying testsets~\citep{elliott-etal-2016-multi30k, elliott-etal-2017-findings, barrault-etal-2018-findings}. Although the systems show increased BLEU scores~\citep{papineni-etal-2002-bleu} compared to the text-only baselines, the benefits have been attributed to regularization~\citep{wu-etal-2021-good}. For example, \citet{elliott-2018-adversarial}, after replacing the relevant images with random ones, did not see the expected drop in performance. This could be caused by the insufficient number of examples in the Multi30k training dataset that require context in the form of the accompanying image for correct translation~\citep{futeral-etal-2023-tackling}.

Earlier works trained models from random initialization~\citep{yin-etal-2020-novel, liu2021gumbel}, but increasingly often researchers utilize pretrained models~\citep{gupta2023cliptrans, futeral-etal-2025-towards, yu-etal-2025-imagination}. 
This transition is motivated by the strong generalization capabilities of the pretrained models, acquired through training on large amounts of data.
The popular models include CLIP~\citep{radford2021learning} and BLIP~\citep{li2022blip} for image feature extraction, and Large Language Models, such as mBART~\citep{liu-etal-2020-multilingual-denoising} or Qwen~\citep{qwen3technicalreport}, for text understanding and generation.


The emergence of strong general-purpose Multimodal Large Language Models (MLLMs), capable of incorporating visual information without architectural modifications, presents a potential shift in the approach to Multimodal Machine Translation~\citep{alayrac2022flamingo, awadalla2023openflamingo, Qwen2.5-VL, chen2024expanding}. While \citet{xiong-zhao-2025-giift} and \citet{yu-etal-2025-imagination} evaluated their methods against MLLMs, and \citet{zhu-etal-2024-multilingual} performed a large scale evaluation of publicly-available MLLMs, we concentrate on instruction-tuning~\citep{wei2022finetuned, 10.1145/3777411, hu2022lora} and evaluating the open-weights models~\citep{laurencon2023obelics, Qwen2.5-VL}.

Our method is inspired by Metric-based Fine-tuning and Loss Weighting, the two separate techniques introduced for encoder-decoder models by \citet{maka-etal-2025-train} for the task of Context-aware MT. Importantly, both techniques used only a textual modality, and the loss weighting was based on the external annotations. In this work, we combine both techniques, and adapt it to the multimodal setting of Image-guided MT. Furthermore, we introduce a new Metric that uses incongruent images instead of not providing an image.



\section{Method}
\label{sec:method}

Our proposed method consists of two steps. First, we identify which tokens in the target sentence require context. To do this, we measure how much the output probabilities of the pretrained model change when the context image is present or not (alternatively, when the image is not related to the text). Second, we use this information during training to improve the visual sensitivity of the model by increasing the loss corresponding to the context-dependent tokens.

Similar to \citet{maka-etal-2025-train}, we employ the Point-wise Cross-Mutual Information (PCXMI) metric~\citet{fernandes-etal-2023-translation} for the identification step. The original PCXMI metric compares the model's output probabilities with and without previous sentences as context. We adapt it to the setting of Image-guided MT by treating the accompanying image as context. The equation below calculates the $PCXMI$ for the $j$-th token in the target sequence:
\begin{equation}\label{eq:pcxmi}
\begin{aligned} 
    PCXMI_j = log \frac{q\left(y_{j}|y_{t<j}, x, C\right)}{q\left(y_{j}|y_{t<j}, x\right)},
\end{aligned}
\end{equation}
where $q\left(y_{j}|y_{t<j}, x, C\right)$ is the probability of the $j$-th token conditioned on the previous tokens $y_{t<j}$, source sentence $x$, and the image $C$, and $q\left(y_{j}|y_{t<j}, x\right)$ is the probability conditioned only on the previous tokens $y_{t<j}$ and the source sentence $x$. Crucially, the model has to be able to perform translation both with and without the image to use this metric.

Additionally, we propose a modification to the metric, where the comparison is between the probabilities with the correct and incorrect (incongruent) images. We refer to this metric as Congruency-based Point-wise Cross-Mutual Information (CPCMXI):
\begin{equation}\label{eq:congr_pcxmi}
\begin{aligned} 
    CPCXMI_{j} = log \frac{q(y_{j}|y_{t<j}, x, C)}{q(y_{j}|y_{t<j}, x, \tilde{C})},
\end{aligned}
\end{equation}
where $q(y_{j}|y_{t<j}, x, \tilde{C})$ is the probability conditioned on the previous tokens $y_{t<j}$, source sentence $x$, and incongruent image $\tilde{C}$, which we randomly sample from the training dataset. The advantage over the standard PCXMI is that it can be used with models that always require the image to be present. 

In practice, both metrics rely on the outputs of the model in different situations and are, therefore, noisy. To mitigate this, we calculate both metrics for each example $d_i$ in the training dataset $\mathcal{D}$, and take the maximum value for each token as the final image dependence metric:
\begin{equation}\label{eq:pcxmi}
\begin{aligned} 
     a_{i,j} = \max\left(PCXMI_{i,j}, CPCXMI_{i,j}\right).
\end{aligned}
\end{equation}
Taking the maximum increases the reliability of the metric in situations where one or the other metric fails to mark a token as context-dependent.

In the second step, we apply token-level weighting of the standard Negative Log Likelihood loss based on the calculated context dependence metric~\citep{maka-etal-2025-train}:
\begin{equation}\label{eq:loss-function}
\begin{aligned} 
    \mathcal{L} = - \frac{1}{|D|} \sum_{(d_{i}, a_i) \sim D} \frac{1}{|y_i|} \sum_{j=1}^{|y_i|} w\left(a_{i, j}\right) log\left(\hat{y}_{i,j}\right),
\end{aligned}
\end{equation}
where $d_i$ is the $i$-th example sampled from the training dataset $\mathcal{D}$ with the calculated metric $a_i$ for each token in the target sentence, $\hat{y}_{i,j}$ is the output probability of the model for the $j$-th token, and $w\left(a_{i, j}\right)$ is the weight function based on the value of the metric $a_{i, j}$ for the $j$-th token. We use the threshold function to obtain the weighting values:
\begin{equation} \label{eq:weight}
\begin{aligned} 
w(a_{i,j}) = \begin{cases}
1 + \alpha, &\text{if ~~} a_{i,j} > \tau, \\
1, &\text{otherwise}, \\
\end{cases}
\end{aligned}
\end{equation}
where the hyper-parameters $\alpha$ and $\tau$ control the loss increase and the threshold, respectively. 

Intuitively, we penalize the model more for the mistakes in tokens that require context. It should be noted that our method relies on the identification of the image-dependent tokens, which is done by the model itself. This means that we bias the model towards the tokens for which it already exhibits strong visual sensitivity. 

\section{Experiments}

We experimentally evaluated the effects of our method on fine-tuning MLLMs on the task of MMT. 
We are specifically interested in observing how it affects the visual sensitivity of the tested models.

We experimented with three popular open-weights instruction-tuned models: Qwen3-VL~4B~Instruct\footnote{\url{https://huggingface.co/Qwen/Qwen3-VL-4B-Instruct}}, Qwen3-VL~8B~Instruct\footnote{\url{https://huggingface.co/Qwen/Qwen3-VL-8B-Instruct}}~\citep{Qwen2.5-VL, qwen3technicalreport}, and IDEFICS 9B Instruct\footnote{\url{https://huggingface.co/HuggingFaceM4/idefics-9b-instruct}}~\citep{laurencon2023obelics}. We used LoRA~\citep{hu2022lora} to fine-tune the models on the Multi30k dataset~\citep{elliott-etal-2016-multi30k}. We experimented with three language directions: English-to-German, English-to-French, and English-to-Czech.

\begin{table*}[ht]
\centering
\tiny
\begin{tabular}{lrrrrrr|rrr} 
\hline
\textbf{Model} & \multicolumn{2}{c}{\textbf{test 2016}} & \multicolumn{2}{c}{\textbf{test 2017}} & \multicolumn{2}{c|}{\textbf{MSCOCO}} & \multicolumn{2}{c}{\textbf{Average}} & \textbf{CoMMuTE} \\ 
~ & \textbf{BLEU} & \textbf{COMET} & \textbf{BLEU} & \textbf{COMET} & \textbf{BLEU} & \textbf{COMET} & \textbf{BLEU} & \textbf{COMET} & \textbf{Accuracy} \\ 
\hline
\textbf{Qwen3-VL 4B} & 38.51 & 85.56 & 32.95 & 84.90 & 29.94 & 81.24 & 33.80 & 83.90 & 74.00\% \\ 
~~~Fine-tuning & \textbf{42.61}$\pm$0.12 & 86.66$\pm$0.06 & 38.08$\pm$0.38 & 85.98$\pm$0.06 & 35.22$\pm$0.45 & 82.70$\pm$0.13 & 38.64$\pm$0.18 & 85.11$\pm$0.05 & 71.67$\pm$0.19\% \\ 
~~~VMLM & 42.52$\pm$0.19 & \underline{86.79}$\pm$0.03 & \textbf{38.48}$\pm$0.43 & \textbf{86.34}$\pm$0.06 & \textbf{36.47}$\pm$0.09 & \underline{82.95}$\pm$0.17 & \textbf{39.16}$\pm$0.06 & \textbf{85.36}$\pm$0.05 & 71.67$\pm$0.38\% \\ 
~~~Metric Fine-tuning & \underline{42.57}$\pm$0.22 & 86.47$\pm$0.13 & \underline{38.43}$\pm$0.10 & 85.99$\pm$0.12 & \underline{36.18}$\pm$0.16 & 82.86$\pm$0.18 & \underline{39.06}$\pm$0.15 & 85.10$\pm$0.11 & 72.89$\pm$0.44\% \\ 
~~~Word Dropout & 42.20$\pm$0.19 & 86.62$\pm$0.06 & 37.17$\pm$0.19 & 85.80$\pm$0.01 & 35.00$\pm$0.22 & 82.69$\pm$0.16 & 38.13$\pm$0.18 & 85.03$\pm$0.04 & \underline{76.89}$\pm$0.44\% \\ 
~~~Degradation & 42.18$\pm$0.16 & 86.68$\pm$0.00 & 37.89$\pm$0.03 & 85.88$\pm$0.02 & 35.66$\pm$0.26 & 82.85$\pm$0.04 & 38.58$\pm$0.14 & 85.14$\pm$0.02 & 73.44$\pm$0.59\% \\ 
~~~Metric Weighting & 41.91$\pm$0.33 & \textbf{86.80}$\pm$0.04 & 37.29$\pm$0.14 & \underline{86.14}$\pm$0.04 & 35.71$\pm$0.19 & \textbf{83.09}$\pm$0.13 & 38.30$\pm$0.15 & \underline{85.34}$\pm$0.07 & \textbf{77.00}$\pm$0.33\% \\
\hline
\textbf{Qwen3-VL 8B} & 38.24 & 85.64 & 34.24 & 85.77 & 31.78 & 82.25 & 34.76 & 84.55 & 75.33\% \\ 
~~~Fine-tuning & 42.87$\pm$0.54 & 86.68$\pm$0.21 & 38.15$\pm$0.29 & 86.35$\pm$0.10 & 35.93$\pm$0.22 & 83.11$\pm$0.32 & 38.98$\pm$0.26 & 85.38$\pm$0.19 & 69.33$\pm$0.19\% \\ 
~~~VMLM & 42.25$\pm$0.92 & 86.84$\pm$0.15 & 37.24$\pm$0.62 & 86.09$\pm$0.10 & 35.28$\pm$1.35 & 83.33$\pm$0.21 & 38.26$\pm$0.96 & 85.42$\pm$0.14 & 69.78$\pm$0.97\% \\ 
~~~Metric Fine-tuning & 43.15$\pm$0.16 & 86.80$\pm$0.10 & \underline{38.19}$\pm$0.41 & 86.45$\pm$0.07 & \underline{36.19}$\pm$0.34 & \underline{83.41}$\pm$0.12 & \underline{39.18}$\pm$0.28 & 85.55$\pm$0.08 & 73.56$\pm$0.62\% \\ 
~~~Word Dropout & \underline{43.34}$\pm$0.25 & \textbf{87.17}$\pm$0.03 & 37.08$\pm$0.18 & 86.26$\pm$0.09 & 35.46$\pm$0.74 & \textbf{83.55}$\pm$0.11 & 38.63$\pm$0.18 & \textbf{85.66}$\pm$0.04 & \underline{76.00}$\pm$0.33\% \\ 
~~~Degradation & \textbf{43.47}$\pm$0.34 & \underline{87.01}$\pm$0.04 & \textbf{38.38}$\pm$0.14 & \textbf{86.55}$\pm$0.07 & \textbf{36.42}$\pm$0.19 & 83.34$\pm$0.16 & \textbf{39.43}$\pm$0.10 & \underline{85.63}$\pm$0.08 & 72.22$\pm$0.29\% \\ 
~~~Metric Weighting & 42.82$\pm$0.13 & 86.86$\pm$0.11 & 37.38$\pm$0.30 & \underline{86.47}$\pm$0.07 & 35.12$\pm$0.31 & 83.31$\pm$0.15 & 38.44$\pm$0.14 & 85.55$\pm$0.06 & \textbf{77.22}$\pm$0.22\% \\
\hline
\textbf{IDEFICS 9B} & 27.07 & 80.04 & 22.51 & 78.99 & 19.35 & 74.38 & 22.98 & 77.80 & 68.67\% \\ 
~~~Fine-tuning & 43.48$\pm$0.72 & 86.62$\pm$0.21 & 37.13$\pm$0.15 & \underline{85.58}$\pm$0.18 & \underline{34.24}$\pm$0.45 & 82.13$\pm$0.16 & 38.28$\pm$0.43 & 84.78$\pm$0.18 & 65.89$\pm$0.73\% \\ 
~~~VMLM & 43.38$\pm$0.24 & 86.65$\pm$0.05 & \textbf{38.17}$\pm$0.15 & \textbf{85.80}$\pm$0.03 & \textbf{35.69}$\pm$0.28 & \textbf{82.39}$\pm$0.05 & \textbf{39.08}$\pm$0.23 & \textbf{84.95}$\pm$0.04 & 64.44$\pm$0.48\% \\ 
~~~Metric Fine-tuning & 42.88$\pm$0.55 & 86.37$\pm$0.12 & 36.63$\pm$0.29 & 85.10$\pm$0.29 & 34.14$\pm$0.83 & 81.88$\pm$0.25 & 37.89$\pm$0.55 & 84.45$\pm$0.22 & 67.89$\pm$0.29\% \\ 
~~~Word Dropout & 43.56$\pm$0.26 & \textbf{86.77}$\pm$0.09 & 36.77$\pm$0.49 & 85.51$\pm$0.16 & 34.17$\pm$0.19 & \underline{82.23}$\pm$0.11 & 38.17$\pm$0.26 & \underline{84.83}$\pm$0.09 & \textbf{70.22}$\pm$0.73\% \\ 
~~~Degradation & \textbf{44.07}$\pm$0.22 & \underline{86.71}$\pm$0.04 & 37.36$\pm$0.11 & 85.44$\pm$0.07 & 34.18$\pm$0.18 & 82.01$\pm$0.04 & \underline{38.54}$\pm$0.07 & 84.72$\pm$0.02 & 67.78$\pm$0.11\% \\ 
~~~Metric Weighting & \underline{43.58}$\pm$0.61 & 86.53$\pm$0.10 & \underline{37.42}$\pm$0.15 & 85.38$\pm$0.10 & 34.12$\pm$0.17 & 81.96$\pm$0.18 & 38.37$\pm$0.19 & 84.62$\pm$0.11 & \underline{69.00}$\pm$0.33\% \\
\hline
\end{tabular}
\caption{Results for \textbf{English-to-German} in terms of BLEU and COMET on the Multi30k testsets (separately and averaged over the testsets) and the accuracy on the CoMMuTE contrastive dataset. The results are averaged over three separate runs, and $\pm$ shows the standard error. Best results for each base model in each column are marked in \textbf{bold}, and the second best are \underline{underlined}.}
\label{tab:results-ende}
\end{table*}

We compared our method with the following baselines:
\begin{itemize}[topsep=3pt,itemsep=3pt,partopsep=0pt, parsep=0pt, itemindent=15pt, leftmargin=0pt]
    \item \textbf{Fine-tuning} - simply fine-tuning the model on the Image-guided MT training dataset.
    \item \textbf{Visually Conditioned Masked Language Modeling (VMLM)} - fine-tuning with the auxiliary task, where the model is presented with the source sentence containing masked tokens, and it has to predict those masked tokens~\citep{futeral-etal-2023-tackling, futeral-etal-2025-towards}. We mask words with the probability $p$, which is a hyper-parameter, with a string \verb|MASK{n}|, where \verb|n| goes from 1 to the total number of masked words in the source text. Because VMLM does not require a parallel corpus, often an additional multimodal dataset in the source language is used. In our setting, we use only the Multi30k dataset, discarding the target-side sentences, to keep the methods using comparable amounts of data.
    \item \textbf{Metric-based Fine-tuning} - we adapt this method from the original text-only version~\citep{maka-etal-2025-train}. After the initial fine-tuning, we calculate the PCXMI and CPCXMI metrics and select top-$k$ examples with the highest maximum values of the calculated metrics for further fine-tuning. We take the maximum of the two metrics, as this yielded better results in the preliminary experiments than either of the metrics on their own.
    \item \textbf{Word Dropout} - dropping random tokens from the source sentence to force the model to use the image to improve translation, adapted to Image-guided Machine Translation from CoWord Dropout~\citep{fernandes-etal-2021-measuring}. The probability of dropping out a token is controlled by the hyper-parameter $p$. In the preliminary experiments, we determined that dropout on the word-level was better than token-level. Therefore, we drop out words based on the Moses Tokenizer\footnote{We used the Sacremoses implementation \url{https://github.com/hplt-project/sacremoses} with default parameters.}\citep{koehn-etal-2007-moses}.
    \item \textbf{Source Degradation} - masking specific words in the source sentence during training to enhance the visual sensitivity of the model~\citep{cheng-etal-2024-soul}. The method first identifies the words in three categories: color, person, and noun. Second, the identified words are replaced with the mask strings \verb|MASK_C|, \verb|MASK_P|, \verb|MASK_N| for colors, people, and nouns, respectively (in the original paper, the special tokens have been used instead of words). We used the annotations for types of tokens in the Multi30k dataset from \citet{li-etal-2022-vision} repository\footnote{\url{https://github.com/libeineu/fairseq_mmt}.}.
\end{itemize}

We first tuned the hyper-parameters of Fine-tuning in the preliminary experiments on the English-to-German language direction and kept them the same for other methods. 
For Metric-based Fine-tuning, we selected the number of examples for the second-stage fine-tuning in the preliminary experiments and tuned the learning rate separately for each language direction and model out of $\{1e-5, 2e-5, 5e-5, 1e-4\}$.
For the VMLM, Word Dropout, and Source Degradation methods, we tuned the probability of masking and dropping out words $p \in  \{0.1, 0.2, 0.3\}$. 
For our method, we first tuned the hyper-parameter $\alpha=8$ in the preliminary experiments on the English-to-German language direction, and tested three values of the $\tau$ hyper-parameter for each language direction and model ($\{5, 6, 7\}$, and $\{1, 1.5, 2\}$ for Qwen3-VL and IDEFICS models, respectively). This is because they depend on the distribution of PCXMI and CPCXMI metrics on the training dataset, which are based on the model's outputs (see Section~\ref{sec:pcxim-distributions}). 
We determined that the appropriate range for the $\tau$ values is between the 80th and 90th percentile of the distribution of maximum metric values in training examples. Further details of the training, including prompts and hyper-parameters can be found in Appendix~\ref{sec:training-details}.

We measured BLEU\footnote{\url{https://github.com/mjpost/sacrebleu}}~\citep{papineni-etal-2002-bleu} and COMET\footnote{We used the \texttt{Unbabel/wmt22-comet-da} model.}~\citep{rei-etal-2020-comet} of the translations on the available Multi30k test sets: \textit{test 2016}, \textit{test 2017}, and \textit{MSCOCO} for English-to-German and English-to-French, and \textit{test 2016} for English-to-Czech~\citep{elliott-etal-2016-multi30k, elliott-etal-2017-findings, barrault-etal-2018-findings}. We used the default generation parameters for the tested models. To evaluate the visual sensitivity of the fine-tuned models, we employ the CoMMuTE contrastive dataset~\citep{futeral-etal-2023-tackling}. In this dataset, the models are presented with an image, an ambiguous source sentence, and two possible translations, where the correct one depends on the accompanying image. The dataset measures the accuracy of the models by assigning higher probabilities to correct target sentences.
Following~\citet{futeral-etal-2023-tackling}, we repeated the experiments three times with different seeds and averaged the results.


\subsection{Results}

\begin{table*}
\centering
\tiny
\begin{tabular}{lrrrrrr|rrr} 
\hline
\textbf{Model} & \multicolumn{2}{c}{\textbf{test 2016}} & \multicolumn{2}{c}{\textbf{test 2017}} & \multicolumn{2}{c|}{\textbf{MSCOCO}} & \multicolumn{2}{c}{\textbf{Average}} & \textbf{CoMMuTE} \\ 
~ & \textbf{BLEU} & \textbf{COMET} & \textbf{BLEU} & \textbf{COMET} & \textbf{BLEU} & \textbf{COMET} & \textbf{BLEU} & \textbf{COMET} & \textbf{Accuracy} \\ 
\hline
\textbf{Qwen3-VL 4B} & 43.77 & 86.04 & 43.47 & 86.31 & 43.24 & 83.69 & 43.49 & 85.35 & 75.00\% \\ 
~~~Fine-tuning & 67.03$\pm$0.36 & \textbf{89.81}$\pm$0.05 & \textbf{61.62}$\pm$0.13 & \underline{89.25}$\pm$0.10 & 50.90$\pm$0.43 & \textbf{85.37}$\pm$0.12 & \underline{59.85}$\pm$0.08 & \textbf{88.14}$\pm$0.03 & 81.17$\pm$0.37\% \\ 
~~~VMLM & \underline{67.37}$\pm$0.25 & \underline{89.71}$\pm$0.05 & 61.00$\pm$0.19 & 89.23$\pm$0.10 & 50.71$\pm$0.57 & 85.10$\pm$0.07 & 59.69$\pm$0.21 & 88.01$\pm$0.05 & 81.71$\pm$0.39\% \\ 
~~~Metric Fine-tuning & 66.83$\pm$0.41 & 89.43$\pm$0.12 & 60.98$\pm$0.11 & 88.97$\pm$0.02 & 49.73$\pm$0.09 & 84.71$\pm$0.15 & 59.18$\pm$0.13 & 87.70$\pm$0.09 & 81.17$\pm$0.19\% \\ 
~~~Word Dropout & 66.46$\pm$0.31 & 89.58$\pm$0.09 & 60.98$\pm$0.05 & 89.10$\pm$0.02 & \underline{51.01}$\pm$0.44 & 85.15$\pm$0.04 & 59.48$\pm$0.27 & 87.94$\pm$0.04 & 84.42$\pm$0.68\% \\ 
~~~Degradation & \textbf{67.42}$\pm$0.18 & 89.69$\pm$0.04 & \underline{61.58}$\pm$0.13 & \textbf{89.26}$\pm$0.06 & \textbf{51.09}$\pm$0.18 & \underline{85.33}$\pm$0.06 & \textbf{60.03}$\pm$0.04 & \underline{88.09}$\pm$0.02 & \underline{84.52}$\pm$0.11\% \\ 
~~~Metric Weighting & 67.22$\pm$0.28 & 89.69$\pm$0.03 & 61.04$\pm$0.23 & 89.15$\pm$0.04 & 50.50$\pm$0.37 & 85.07$\pm$0.03 & 59.59$\pm$0.22 & 87.97$\pm$0.03 & \textbf{84.74}$\pm$0.50\% \\
\hline
\textbf{Qwen3-VL 8B} & 47.53 & 86.93 & 47.32 & 87.14 & 46.40 & 84.36 & 47.08 & 86.14 & 75.00\% \\ 
~~~Fine-tuning & \underline{68.56}$\pm$0.17 & \underline{90.06}$\pm$0.09 & 62.30$\pm$0.15 & \underline{89.50}$\pm$0.05 & \textbf{52.66}$\pm$0.31 & \textbf{85.74}$\pm$0.17 & \underline{61.17}$\pm$0.13 & \textbf{88.43}$\pm$0.06 & 83.12$\pm$0.00\% \\ 
~~~VMLM & 68.15$\pm$0.12 & 90.01$\pm$0.09 & 61.52$\pm$0.08 & 89.48$\pm$0.11 & \underline{51.56}$\pm$0.20 & 85.57$\pm$0.08 & 60.41$\pm$0.05 & 88.35$\pm$0.01 & 81.49$\pm$0.19\% \\ 
~~~Metric Fine-tuning & \textbf{69.32}$\pm$0.10 & \textbf{90.07}$\pm$0.04 & \textbf{62.91}$\pm$0.27 & \textbf{89.55}$\pm$0.03 & 51.31$\pm$0.35 & \underline{85.60}$\pm$0.10 & \textbf{61.18}$\pm$0.02 & \underline{88.41}$\pm$0.05 & 84.42$\pm$0.50\% \\ 
~~~Word Dropout & 67.20$\pm$0.18 & 89.81$\pm$0.05 & 61.14$\pm$0.52 & 89.03$\pm$0.09 & 51.27$\pm$0.15 & 85.23$\pm$0.07 & 59.87$\pm$0.09 & 88.02$\pm$0.04 & \textbf{85.28}$\pm$0.57\% \\ 
~~~Degradation & 67.29$\pm$0.09 & 89.93$\pm$0.04 & 62.07$\pm$0.15 & 89.29$\pm$0.07 & 50.99$\pm$0.18 & 85.44$\pm$0.09 & 60.12$\pm$0.02 & 88.22$\pm$0.06 & 81.28$\pm$0.29\% \\ 
~~~Metric Weighting & 67.74$\pm$0.16 & 89.98$\pm$0.06 & \underline{62.68}$\pm$0.19 & 89.37$\pm$0.03 & 51.19$\pm$0.18 & 85.56$\pm$0.16 & 60.54$\pm$0.07 & 88.30$\pm$0.04 & \underline{85.06}$\pm$0.50\% \\
\hline
\textbf{IDEFICS 9B} & 33.35 & 82.26 & 32.15 & 81.76 & 31.80 & 78.64 & 32.43 & 80.89 & 71.75\% \\ 
~~~Fine-tuning & \underline{67.00}$\pm$0.19 & \underline{89.39}$\pm$0.11 & 60.39$\pm$0.13 & \underline{88.62}$\pm$0.09 & \underline{50.58}$\pm$0.24 & 84.87$\pm$0.09 & 59.32$\pm$0.12 & 87.63$\pm$0.03 & 76.73$\pm$0.22\% \\ 
~~~VMLM & 66.78$\pm$0.04 & \textbf{89.47}$\pm$0.05 & \underline{60.44}$\pm$0.05 & \textbf{88.68}$\pm$0.11 & \textbf{50.83}$\pm$0.36 & \underline{85.03}$\pm$0.15 & \underline{59.35}$\pm$0.12 & \textbf{87.73}$\pm$0.02 & 78.57$\pm$0.37\% \\ 
~~~Metric Fine-tuning & \textbf{67.52}$\pm$0.24 & \textbf{89.47}$\pm$0.16 & 60.19$\pm$0.18 & 88.54$\pm$0.02 & 50.41$\pm$0.38 & 84.94$\pm$0.16 & \textbf{59.37}$\pm$0.25 & 87.65$\pm$0.11 & 77.92$\pm$0.32\% \\ 
~~~Word Dropout & 65.82$\pm$0.12 & 89.19$\pm$0.06 & 59.74$\pm$0.11 & 88.43$\pm$0.03 & 49.82$\pm$0.24 & 84.80$\pm$0.19 & 58.46$\pm$0.08 & 87.47$\pm$0.05 & \underline{78.68}$\pm$0.43\% \\ 
~~~Degradation & \underline{67.00}$\pm$0.13 & 89.28$\pm$0.03 & \textbf{60.50}$\pm$0.07 & 88.54$\pm$0.01 & 50.08$\pm$0.33 & 84.96$\pm$0.13 & 59.19$\pm$0.09 & 87.59$\pm$0.05 & 78.14$\pm$0.89\% \\ 
~~~Metric Weighting & 66.55$\pm$0.12 & 89.22$\pm$0.05 & 60.29$\pm$0.07 & 88.48$\pm$0.08 & 50.51$\pm$0.22 & \textbf{85.35}$\pm$0.08 & 59.12$\pm$0.05 & \underline{87.69}$\pm$0.05 & \textbf{78.79}$\pm$0.29\% \\
\hline
\end{tabular}
\caption{Results for \textbf{English-to-French} in terms of BLEU and COMET on the Multi30k testsets (separately and averaged over the testsets) and the accuracy on the CoMMuTE contrastive dataset. The results are averaged over three separate runs, and $\pm$ shows the standard error. Best results for each base model in each column are marked in \textbf{bold}, and the second best are \underline{underlined}.}
\label{tab:results-enfr}
\end{table*}

\begin{table}[ht]
\centering
\tiny
\begin{tabular}{lrrr} 
\hline
\textbf{Model} & \multicolumn{2}{c}{\textbf{test 2016}} & \textbf{CoMMuTE} \\ 
~ & \textbf{BLEU} & \textbf{COMET} & \textbf{Accuracy} \\ 
\hline
\textbf{Qwen3-VL 4B} & 23.12 & 83.43 & 67.21\% \\ 
~~~Fine-tuning & \underline{35.82}$\pm$0.25 & \textbf{88.63}$\pm$0.05 & 74.35$\pm$0.37\% \\ 
~~~VMLM & 35.66$\pm$0.08 & 88.46$\pm$0.10 & 74.68$\pm$0.50\% \\ 
~~~Metric Fine-tuning & 35.19$\pm$0.16 & 88.39$\pm$0.06 & \underline{75.54}$\pm$1.07\% \\ 
~~~Word Dropout & 35.72$\pm$0.20 & \underline{88.62}$\pm$0.09 & 75.43$\pm$0.66\% \\ 
~~~Degradation & \textbf{36.19}$\pm$0.40 & \underline{88.62}$\pm$0.02 & 74.89$\pm$0.29\% \\ 
~~~Metric Weighting & \textbf{36.19}$\pm$0.31 & 88.46$\pm$0.12 & \textbf{76.41}$\pm$0.43\% \\ 
\hline
\textbf{Qwen3-VL 8B} & 26.82 & 86.37 & 74.03\% \\ 
~~~Fine-tuning & 36.17$\pm$0.08 & \textbf{89.06}$\pm$0.06 & 78.03$\pm$0.22\% \\ 
~~~VMLM & \textbf{36.99}$\pm$0.34 & 88.84$\pm$0.13 & 78.46$\pm$0.29\% \\ 
~~~Metric Fine-tuning & 36.42$\pm$0.15 & 88.64$\pm$0.03 & 79.98$\pm$0.22\% \\ 
~~~Word Dropout & 35.98$\pm$0.29 & \underline{88.91}$\pm$0.12 & \textbf{80.84}$\pm$0.19\% \\ 
~~~Degradation & \underline{36.68}$\pm$0.52 & 88.87$\pm$0.12 & 78.57$\pm$0.19\% \\ 
~~~Metric Weighting & 36.52$\pm$0.24 & 88.81$\pm$0.03 & \underline{80.19}$\pm$0.32\% \\ 
\hline
\textbf{IDEFICS 9B} & 10.29 & 70.42 & 64.29\% \\ 
~~~Fine-tuning & \underline{36.80}$\pm$0.18 & 88.27$\pm$0.04 & 68.72$\pm$0.76\% \\ 
~~~VMLM & 36.41$\pm$0.16 & \underline{88.35}$\pm$0.11 & 70.02$\pm$0.47\% \\ 
~~~Metric Fine-tuning & 36.33$\pm$0.33 & 88.16$\pm$0.09 & 71.43$\pm$0.32\% \\ 
~~~Word Dropout & \textbf{36.89}$\pm$0.06 & \textbf{88.65}$\pm$0.09 & \underline{73.27}$\pm$0.66\% \\ 
~~~Degradation & 36.66$\pm$0.05 & 88.28$\pm$0.09 & 72.84$\pm$0.92\% \\ 
~~~Metric Weighting & 36.76$\pm$0.18 & 88.15$\pm$0.08 & \textbf{74.03}$\pm$0.32\% \\ 
\hline
\end{tabular}
\caption{Results for \textbf{English-to-Czech} in terms of BLEU and COMET on the Multi30k test 2016 testset and the accuracy on the CoMMuTE contrastive dataset. The results are averaged over three separate runs, and $\pm$ shows the standard error. Best results for each base model in each column are marked in \textbf{bold}, and the second best are \underline{underlined}.}
\label{tab:results-encs}
\end{table}

The results in terms of BLEU and COMET on the \textit{test 2016}, \textit{test 2017}, and \textit{MSCOCO} testsets, and CoMMuTE accuracy are presented in Tables~\ref{tab:results-ende}~and~\ref{tab:results-enfr} for the English-to-German and English-to-French language directions. To improve the readability of the results, we additionally included the BLEU and COMET averaged over the testsets. The results in terms of BLEU and COMET on the \textit{test 2016} testset and CoMMuTE accuracy for the English-to-Czech language direction are shown in Table~\ref{tab:results-encs}. 
Results are averaged over three runs for each method, and standard errors are reported in the tables.

In terms of BLEU and COMET, Qwen3-VL 8B exhibited better performance than the other two models, with IDEFICX 9B scoring lower than the smaller Qwen-VL 4B model on all language directions. While this could be caused by different generation method (no sampling for the IDEFICX model), this gap is removed after fine-tuning on the Multi30k dataset. The CoMMuTE accuracy of the baseline models is similar between English-to-German and English-to-French, and there is a clear drop for English-to-Czech for Qwen-3-VL 4B and IDEFICS 9B models (more than 6 and 4 percentage points, respectively). This could be attributed to lower amounts of English-to-Czech parallel and Czech monolingual data in the training corpora.

For English-to-German, fine-tuning reduces the visual grounding, as measured by the CoMMuTE accuracy, compared to the baseline model. Only Word Dropout and Metric-based Loss Weighting retrieve and surpass the performance of the baseline model. This effect is not visible for the two remaining language directions, where each technique improves the baseline performance. 
We hypothesize that the drop in CoMMuTE accuracy from English to German stems from linguistic proximity (both are Germanic languages). This proximity may make the translation task easier for the models to learn during fine-tuning, reducing their reliance on the visual modality to resolve ambiguity. We leave evaluating this hypothesis for future work. VMLM did not yield large improvements in visual sensitivity compared to other methods. We attribute this to our use of only a parallel multimodal dataset for this task. We expect it to show larger improvements when trained on a large-scale monolingual multimodal dataset. However, training it already used twice as much computational resources compared to other methods (see Appendix~\ref{sec:training-details}), and scaling it further falls outside the scope of this work.

\begin{figure*}[h]
\centering
\small
\begin{tabular}{ll}
\multirowcell{6}[0pt][r]{\includegraphics[height=2.6cm]{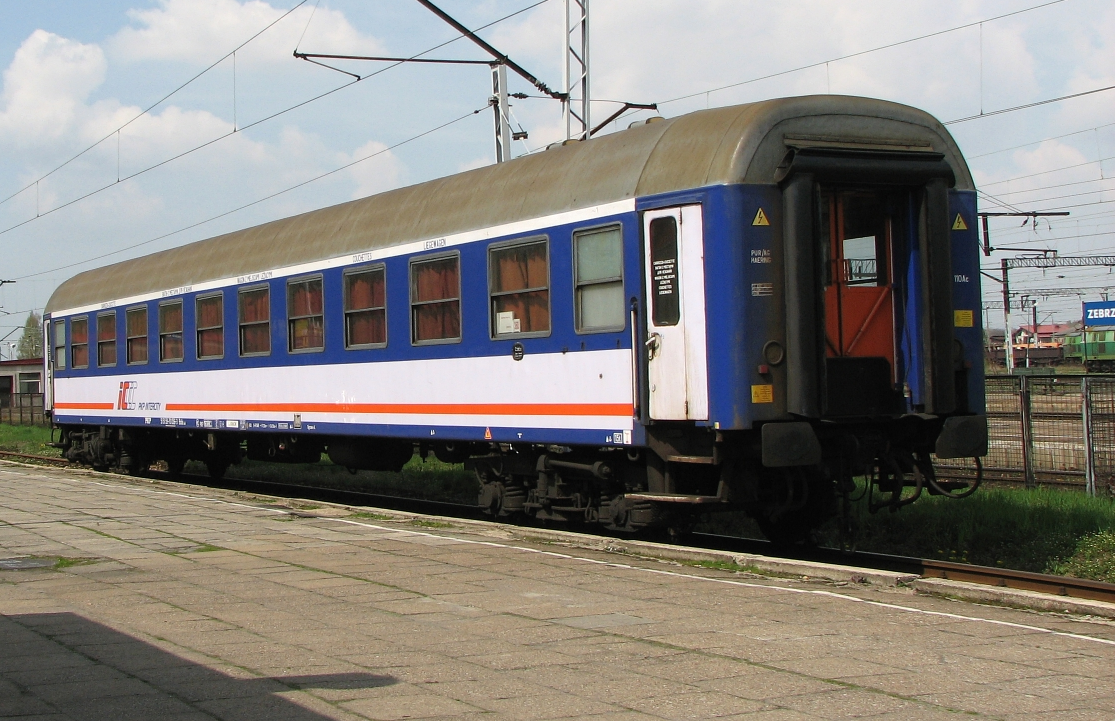}} & English: We changed \incorrect{coaches}. \\
~ & German: Wir haben den \incorrect{Waggon} gewechselt. \\
~ & \multirowcell{3}[0pt][l]{\includegraphics[width=9.2cm]{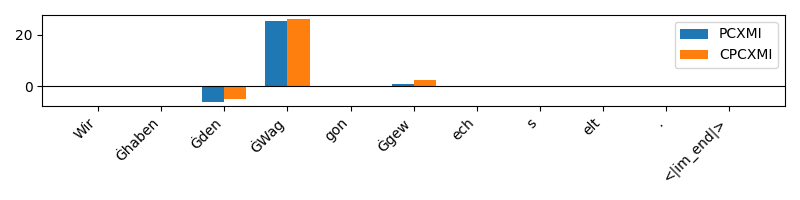}} \\
~ & ~ \\
~ & ~ \\
~ & ~ \\
~ & ~ \\
~ & ~ \\
\multirowcell{6}[0pt][r]{\includegraphics[height=2.6cm]{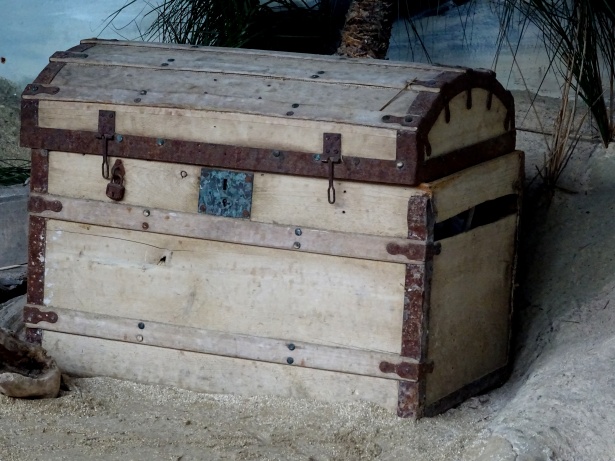}} & English: My \incorrect{chest} is quite heavy... \\
~ & German: Mein \incorrect{Koffer} ist ziemlich schwer. \\
~ & \multirowcell{3}[0pt][l]{\includegraphics[width=9cm]{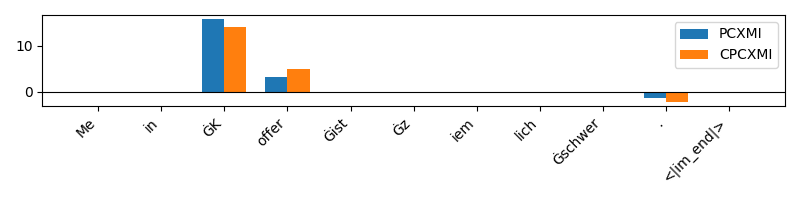}} \\
~ & ~ \\
~ & ~ \\
~ & ~ \\
~ & ~ \\
\end{tabular}
\caption{Two examples from the English-to-German CoMMuTE dataset with the per-token values of PCXMI and CPCXMI metrics for the Qwen3-VL 4B model. In \incorrect{red} we mark the ambiguous words in the source side and the corresponding words in the target side.}
\label{tab:pcxmi-examples}
\end{figure*}


While the highest results of CoMMuTE accuracy are significantly different from the baseline models and Fine-tuning (according to Welch's t-test), due to the low number of runs, the results for the top methods (marked as bold and underlined in the result tables) are not significantly different. Nevertheless, Metric-based Loss Weighting achieved the highest results in terms of CoMMuTE accuracy on 6 out of 9 experimental settings, and achieved the second-best result for the remaining three. Word Dropout showed the best result in three settings and was the second best in four of them. Source Degradation achieved the second best result on English-to-French with the Qwen3-VL 4B model, and Metric-based Fine-tuning on English-to-Czech with the same model. Notably, the high score on CoMMuTE does not correlate well with the achieved BLEU and COMET scores, indicating that the general translation quality and visual grounding are not necessarily mutual, which corroborates the findings of~\citet{futeral-etal-2025-towards}. This can be attributed to the examples in the Multi30k testsets not requiring images for good translation~\citep{caglayan-etal-2019-probing, wu-etal-2021-good}. We further investigate this phenomenon in Section~\ref{sec:qualitatie-analysis}, where we qualitatively analyze the selected translations in the MSCOCO English-to-French dataset.

\section{Analysis}

In this section, we take a deeper look at the PCXMI and CPCXMI distributions. Next, we examine translation examples produced by our method and standard fine-tuning. Lastly, we experiment with combining our method with Word Dropout, another technique that achieves strong results. Additionally, in Appendix~\ref{sec:ablation} we present the ablation study.

\subsection{Metrics Analysis}
\label{sec:pcxim-distributions}

In Figure~\ref{tab:pcxmi-examples}, we present two examples of the per-token values of the PCXMI and CPCXMI metrics. Further examples can be found in Appendix~\ref{sec:metrics-analysis-extended}. While both metrics are capable of identifying the tokens that require image for disambiguation, they are not perfect, and can ignore or misidentify those tokens. This can be partially mitigated by taking the maximum of both metrics.

In Appendix~\ref{sec:metrics-analysis-extended}, we present the distributions of the calculated metrics (PCXMI, CPCXMI, and their maximum) on the training subset of the Multi30k dataset for each tested model and language direction. The values of the metrics depend heavily on the model, as can be seen in the difference between both the Qwen3-VL models and IDEFICS 9B. The difference between language directions for the same model is a lot smaller. Additionally, all distributions show a concentration of values close to zero and a long tail at high values. This means that most examples in the dataset do not require, according to the tested models, images to disambiguate. The distributions for Multi30k and CoMMute datasets are compared in Appendix~\ref{sec:metrics-analysis-extended}, where we show that the distribution for the latter is denser for higher values of the metric.

\subsection{Qualitative Analysis}
\label{sec:qualitatie-analysis}

\begin{figure}[ht]
    \centering
    \includegraphics[width=0.9\linewidth]{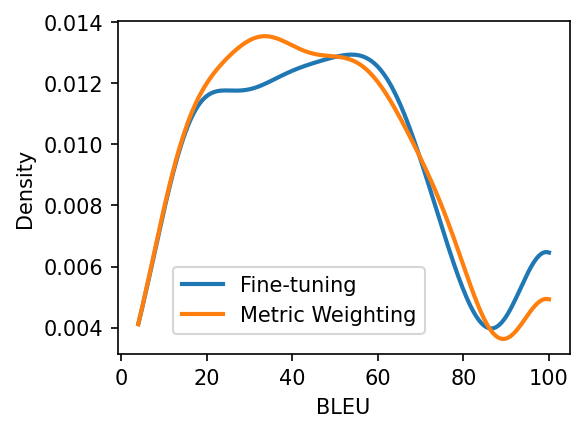}
    \caption{Distributions of BLEU scores of Fine-tuning and Metric-based Loss Weighting on the MSCOCO testset for the English-to-French language direction.}
    \label{fig:bleu-distribution}
\end{figure}

In this section, we investigate the drop in BLEU exhibited by Metric-based Loss Weighting compared to Fine-tuning seen for Qwen3-VL 8B on the English-to-French language direction. We perform the qualitative analysis of the translations of the MSCOCO testset, where the difference in BLEU between the methods is the largest. In Figure~\ref{fig:bleu-distribution}, we present the distributions of per-example BLEU scores for Fine-tuning and our method. They differ mostly in two regions: (1) BLEU scores around 100, where Fine-tuning has more examples, (2) BLEU scores around 30, where the Weighting distribution is denser.

We examined the examples that exhibit the highest difference in BLEU between Fine-tuning and our method (both ways, with both Fine-tuning and our method achieving better scores). We found that the largest drop in example-level BLEU scores mostly corresponds to acceptable translations that differ from the gold target translation. More examples with our descriptions and further details of the analysis are presented in Appendix~\ref{sec:extended-qualitative-analysis}.

\subsection{Combining Weighting with Dropout}

The main results of the tested methods show that Metric-based Loss Weighting achieves the highest CoMMuTE accuracy. The second best performing method, Word Dropout, modifies the source side and is technically not related to our method, which changes the training signal corresponding to the target side. In this section, we test whether the combination of both methods could increase the visual sensitivity of the resulting models even further.

For each example, the Metric-based Loss Weighting is applied (as described in Section~\ref{sec:method}). In cases where there is no token with the metric value greater than the threshold parameter $\tau$, we apply Word Dropout to the source sentence with the specified per-token probability $p$. We test the combined method with the Qwen3-VL 4B model on all three tested language directions (the experimental configuration is the same as before). The results can be seen in Table~\ref{tab:combined}, where we compare with Fine-tuning, and both methods (the full non-aggregated results for English-to-German and English-to-French can be found in Appendix~\ref{sec:extended-results}). The combined method achieves higher CoMMuTE accuracy on English-to-German and English-to-Czech, but did not improve the Metric-based Loss Weighting on English-to-French. Furthermore, the combined method shows reduced BLEU for all tested languages, but the reduction in COMET was not observed. This shows that both methods can work in tandem, increasing the visual grounding even further, but resulting in the distributions of outputs moving away from the gold translations.   

\begin{table}[ht]
\centering
\tiny
\begin{tabular}{lrrr} 
\hline
\textbf{Model} & \multicolumn{2}{c}{\textbf{Average}} & \textbf{CoMMuTE} \\ 
~ & \textbf{BLEU} & \textbf{COMET} & \textbf{Accuracy} \\ 
\hline
\multicolumn{4}{c}{\textbf{English-to-German}} \\
\hline
Qwen3-VL 4B & 33.80 & 83.90 & 74.00\% \\ 
~~~Fine-tuning & \textbf{38.64}$\pm$0.18 & 85.11$\pm$0.05 & 71.67$\pm$0.19\% \\ 
~~~Word Dropout & 38.13$\pm$0.18 & 85.03$\pm$0.04 & 76.89$\pm$0.44\% \\ 
~~~Metric Weighting & \underline{38.30}$\pm$0.15 & \textbf{85.34}$\pm$0.07 & \underline{77.00}$\pm$0.33\% \\ 
~~~Weighting + Dropout & 37.80$\pm$0.11 & \underline{85.23}$\pm$0.09 & \textbf{78.78}$\pm$0.11\% \\ 
\hline
\multicolumn{4}{c}{\textbf{English-to-French}} \\
\hline
Qwen3-VL 4B & 43.49 & 85.35 & 75.00\% \\ 
~~~Fine-tuning & \textbf{59.85}$\pm$0.08 & \textbf{88.14}$\pm$0.03 & 81.17$\pm$0.37\% \\ 
~~~Word Dropout & 59.48$\pm$0.27 & 87.94$\pm$0.04 & 84.42$\pm$0.68\% \\ 
~~~Metric Weighting & \underline{59.59}$\pm$0.22 & \underline{87.97}$\pm$0.03 & \textbf{84.74}$\pm$0.50\% \\ 
~~~Weighting + Dropout & 58.71$\pm$0.22 & 87.93$\pm$0.09 & \underline{84.63}$\pm$0.39\% \\ 
\hline
\multicolumn{4}{c}{\textbf{English-to-Czech}} \\
\hline
Qwen3-VL 4B & 23.12 & 83.43 & 67.21\% \\ 
~~~Fine-tuning & \underline{35.82}$\pm$0.25 & \underline{88.63}$\pm$0.05 & 74.35$\pm$0.37\% \\ 
~~~Word Dropout & 35.72$\pm$0.20 & 88.62$\pm$0.09 & 75.43$\pm$0.66\% \\ 
~~~Metric Weighting & \textbf{36.19}$\pm$0.31 & 88.46$\pm$0.12 & \underline{76.41}$\pm$0.43\% \\ 
~~~Weighting + Dropout & 35.27$\pm$0.39 & \textbf{88.69}$\pm$0.11 & \textbf{77.27}$\pm$0.37\% \\ 
\hline
\end{tabular}
\caption{Results for combined Metric-based Loss Weighting and Word Dropout in terms of BLEU and COMET averaged over the test sets (and three separate runs) for English-to-German and English-to-French, and for \textit{test 2016} for English-to-Czech, and CoMMuTE accuracy.}
\label{tab:combined}
\end{table}

\section{Conclusions}

This work introduced a training strategy for fine-tuning MLLMs for the task of MMT that increases the visual grounding during translation. It increases the loss for the tokens that benefit from context in the form of the image, as measured by the Point-wise Cross-Mutual Information (PCXMI) and Congruency-based PCXMI (CPCXMI). Both metrics rely on the model's output probabilities with and without the image (or with incongruent image in case of CPCXMI).

We empirically tested our technique against existing methods adapted for the task of fine-tuning MLLMs on three models and three language directions, and showed its increased performance in terms of CoMMuTE accuracy, while maintaining strong overall translation quality, as measured by BLEU and COMET. Our results confirm that general-purpose translation metrics (BLEU and COMET) fail to capture how strongly models rely on visual modality for MMT. In future work, we will explore how the metrics can inform masking and word dropout on the source side.

\section{Limitations}
While we investigate three language directions and three sub-families, all of them come from the Indo-European family, and all use the Latin script. This limitation was imposed by the language directions covered by both Multi30k and CoMMuTE datasets. We suspect that the measured performance of the methods goes beyond the tested language pairs, but this claim has not been tested experimentally. Due to the low number of repeated runs and the variability of the measured metrics, the results of the top methods are not different with statistical significance. We limited our investigation to only the same multimodal parallel dataset for each method, while VMLM could use additional multimodal monolingual data.

For Metric-based Loss Weighting, we calculate the image dependence metrics using the baseline models. Updating the metrics during training or using a separate (presumably more visually sensitive) model could yield different results, but we did not test this setting. Furthermore, we limit our method to PCXMI and CPCXMI, and a broader exploration of image dependence metrics could lead to increased performance. Finally, while we tested the methods on three MLLMs, their sizes were limited to less than 10B parameters.

\bibliography{custom,anthology-1,anthology-2}

\appendix

\section{Training Details}
\label{sec:training-details}

We trained the models using the AdamW optimizer~\citep{loshchilov2018decoupled}. We used a single NVIDIA GeForce RTX 3090 24GB for Qwen3-VL 4B, and a single NVIDIA A100 40GB for Qwen3-VL 8B and IDEFICS 9B. The average training times per epoch for each model are presented in Table~\ref{tab:training-times}. The time for Metric-based Fine-tuning is shown only for the second stage training (training on the top-k examples). The training time for VMLM is roughly doubled compared to fine-tuning, because of the auxiliary task of masked language modeling. The times for other methods are comparable, meaning that additional pre-processing (dropout, masking) and post-processing (loss weighting) do not meaningfully increase the training time compared to basic fine-tuning. The training was performed on the Multi30k dataset and evaluated on its testsets and the CoMMuTE contrastive dataset. The sizes of the used datasets can be found in Table~\ref{tab:dataset-sizes}.

\begin{table}[ht]
\centering
\begin{tabular}{lrrr} 
\hline
\textbf{Model} & \textbf{Epoch Time} \\ 
\hline
\textbf{Qwen3-VL 4B} & ~ \\ 
~~~Fine-tuning & 1:43 \\ 
~~~VMLM & 3:30 \\ 
~~~Metric Fine-tuning* & 0:20 \\ 
~~~Word Dropout & 1:43 \\ 
~~~Degradation & 1:43 \\ 
~~~Metric Weighting & 1:43 \\ 
\hline
\textbf{Qwen3-VL 8B} & ~  \\ 
~~~Fine-tuning & 1:16 \\ 
~~~VMLM & 2:24 \\ 
~~~Metric Fine-tuning* & 0:13 \\ 
~~~Word Dropout & 1:14 \\ 
~~~Degradation & 1:15 \\ 
~~~Metric Weighting & 1:14 \\ 
\hline
\textbf{IDEFICS 9B} & ~  \\ 
~~~Fine-tuning & 0:51 \\ 
~~~VMLM & 1:40 \\ 
~~~Metric Fine-tuning* & 0:09 \\ 
~~~Word Dropout & 0:49 \\ 
~~~Degradation & 0:50 \\ 
~~~Metric Weighting & 0:51 \\ 
\hline
\end{tabular}
\caption{Average training times per epoch for the tested models (in hours:minutes). *The time for Metric-based Fine-tuning is only for the second stage of the training.}
\label{tab:training-times}
\end{table}

\begin{table}[ht]
\centering
\begin{tabular}{lrrr} 
\hline
\textbf{Dataset} & \textbf{\# Examples} \\ 
\hline
\textbf{Multi30k} & ~ \\ 
~~~Train & 29,000 \\ 
~~~Validation & 1014 \\ 
~~~Test 2016 & 1000 \\ 
~~~Test 2017 & 1000 \\ 
~~~MSCOCO & 461 \\ 
\hline
\textbf{CoMMuTE} & ~  \\ 
~~~Test & 300 \\ 
~~~Validation & 60 \\ 
\hline
\end{tabular}
\caption{The sizes of the used datasets in terms the number of examples.}
\label{tab:dataset-sizes}
\end{table}

The shared hyper-parameters for all methods and models can be found in Table~\ref{tab:hyper-parameters}. 
For generation, we used the default parameters of all models: multinomial sampling with temperature 0.7 for Qwen3-VL models, and greedy decoding for IDEFICS 9B. The translation prompt can be seen in Listing~\ref{lst:translation-prompt}, and the VMLM prompt in Listing~\ref{lst:vmlm-prompt}.

\begin{code}[caption={The translation prompt used for generation, contrastive evaluation on CoMMuTE, and metric calculation. The source and target languages (in natural language, e.g., "English" or "German") are represented as \texttt{\{src\_lang\}} and \texttt{\{tgt\_lang\}}, respectively. The source sentence is represented as \texttt{\{src\}}. The accompanying image is passed as the input alongside the prompt.}, label={lst:translation-prompt}]
"Translate the following {src_lang} text to {tgt_lang}.\n
{src_lang}: {src}\n
{tgt_lang}:"
\end{code}

\begin{code}[caption={The Visually Conditioned Masked Language Modeling (VMLM) prompt and response used for training. The source language (in natural language, e.g., "English") and masked source text are represented as \texttt{\{src\_lang\}} and \texttt{\{masked\_text\}}, respectively. \texttt{\{masked\_word1\}} and \texttt{\{masked\_word2\}} represent the masked words (the number of masked words can be different between examples). The accompanying image is passed as the input alongside the prompt.}, label={lst:vmlm-prompt}]
Prompt:
"The following {src_lang} sentence has some words replaced with mask tokens.\n
Recover the masked words.\n
Sentence: {masked_text}"

Response: 
"MASK1: {masked_word1}, MASK2:{masked_word2}, ..."
\end{code}

We tuned the specific hyper-parameters for each method:
\begin{itemize}[topsep=3pt,itemsep=3pt,partopsep=0pt, parsep=0pt, itemindent=15pt, leftmargin=0pt]
    \item \textbf{Visually Conditioned Masked Language Modeling (VMLM)} - token masking probability $p\in\{0.1, 0.2, 0.3\}$;
    \item \textbf{Metric-based Fine-tuning} - learning rate values $\{1e-5, 2e-5, 5e-5, 1e-4\}$ with the constant number of top-k examples for the second stage fine-tuning $k=5000$;
    \item \textbf{Word Dropout} - token dropout probability $p\in\{0.1, 0.2, 0.3\}$;
    \item \textbf{Source Degradation} - token masking probability $p\in\{0.1, 0.2, 0.3\}$ for each eligible word;
    \item \textbf{Metric-based Loss Weighting} - threshold $\tau\in\{5, 6, 7\}$ for Qwen3-VL models and $\tau\in\{1, 1.5, 2\}$ for IDEFICS 9B (based on the PCXMI and CPCXMI distributions) with constant loss weighting $\alpha=8$.
\end{itemize}
The tuning was performed based on the validation subsets of the Multi30k and CoMMuTE datasets. First, we took only the runs (averaged over 3 seeds) with the highest BLEU on the Multi30k validation (the values not lower than 0.5 BLEU below the highest for a particular model and language direction). Next, we selected the run with the highest accuracy on the CoMMuTE validation subset. While the validation dataset for Mult30k is available for all tested language directions, CoMMuTE only provides it for English-to-French. For this reason, we translated the target side (French) to both German and Czech using Gemini 3.5 Flash\cprotect\footnote{\verb|gemini-3.5-flash| accessed on 10.07.2026} through Google API. The prompt for translation is presented in Listing~\ref{lst:commute-translate-prompt}.
The found best hyper-parameters for each method and language direction can be found in Table~\ref{tab:best-hyper-parameters}.

\begin{code}[caption={The prompt used to translate the CoMMuTE validation set from French to German and Czech. The source and target expressions in the original dataset are represented as \texttt{\{src\}} and \texttt{\{tgt\}}, respectively. The translation language (German and Czech) is marked as \texttt{\{trans\_lang\}}. The accompanying image is passed as the input alongside the prompt.}, label={lst:commute-translate-prompt}]
"Translate the following English expression to {trans_lang}: {src}.\n
Use the image and the following French translation as references: {tgt}.\n
Output only the translation."
\end{code}

\begin{table}[h!]
    \centering
    \begin{tabular}{lr}
    \hline
    \textbf{Hyper-parameter} & \textbf{Value} \\
    \hline
    Optimizer & AdamW  \\
    Learning Rate & 0.0001* \\
    LR Scheduler & Cosine \\
    LR Warmup Steps & 500 \\
    Weight Decay & 0.01 \\
    Max Grad Norm & 1.0 \\
    Batch Size & 8  \\
    Gradient Accumulation Steps & 8 \\
    Num Epoch & 10 \\
    Patience & 2 \\
    Precision & bf16  \\
    Lora $\alpha$ & 64 \\
    Lora $r$ & 32 \\
    Lora Dropout & 0.05 \\
    Max Length & 200 \\
    Seeds & 1, 2, 3 \\
    \hline
    \end{tabular}
    \caption{The shared hyper-parameters of all tested fine-tuning methods. *We tuned the Learning Rate hyper-parameter for the second stage of the Metric-based Fine-tuning method.}
    \label{tab:hyper-parameters}
\end{table}

\begin{table}
\centering
\footnotesize
\begin{tabular}{lrrr} 
\hline
\textbf{Model} & \textbf{En-De} & \textbf{En-Fr} & \textbf{En-Cs} \\ 
\hline
\textbf{Qwen3-VL 4B} & ~ & ~ & ~ \\ 
~~~VMLM (p) & 0.3 & 0.1 & 0.1 \\ 
~~~Metric Fine-tuning (lr) & 1e-4 & 1e-4 & 1e-5 \\ 
~~~Word Dropout (p) & 0.1 & 0.1 & 0.1 \\ 
~~~Degradation (p) & 0.3 & 0.2 & 0.1 \\ 
~~~Metric Weighting ($\tau$) & 5 & 6 & 6 \\ 
\hline
\textbf{Qwen3-VL 8B} & ~ & ~ & ~ \\ 
~~~VMLM (p) & 0.1 & 0.1 & 0.3 \\ 
~~~Metric Fine-tuning (lr) & 1e-5 & 2e-5 & 5e-5 \\ 
~~~Word Dropout (p) & 0.1 & 0.1 & 0.1 \\ 
~~~Degradation (p) & 0.3 & 0.3 & 0.2 \\ 
~~~Metric Weighting ($\tau$) & 5 & 6 & 5 \\ 
\hline
\textbf{IDEFICS 9B} & ~ & ~ & ~ \\ 
~~~VMLM (p) & 0.1 & 0.2 & 0.3 \\ 
~~~Metric Fine-tuning (lr) & 1e-5 & 2e-5 & 2e-5 \\ 
~~~Word Dropout (p) & 0.1 & 0.1 & 0.1 \\ 
~~~Degradation (p) & 0.2 & 0.2 & 0.3 \\ 
~~~Metric Weighting ($\tau$) & 1.5 & 2 & 1.5 \\ 
\hline
\end{tabular}
\caption{Found best hyper-parameters for each method and language direction. For VMLM, Word Dropout and Degradation, the hyper-parameter is the probability of masking or dropout p. For Metric-based Fine-tuning, the hyper-parameter is the learning rate (lr) of the second stage training. For Metric-based Weighting, the hyper-parameter is the threshold $\tau$.}
\label{tab:best-hyper-parameters}
\end{table}

\section{Ablation Study}
\label{sec:ablation}

\begin{figure*}[ht]
    \centering
    \includegraphics[width=1.0\linewidth]{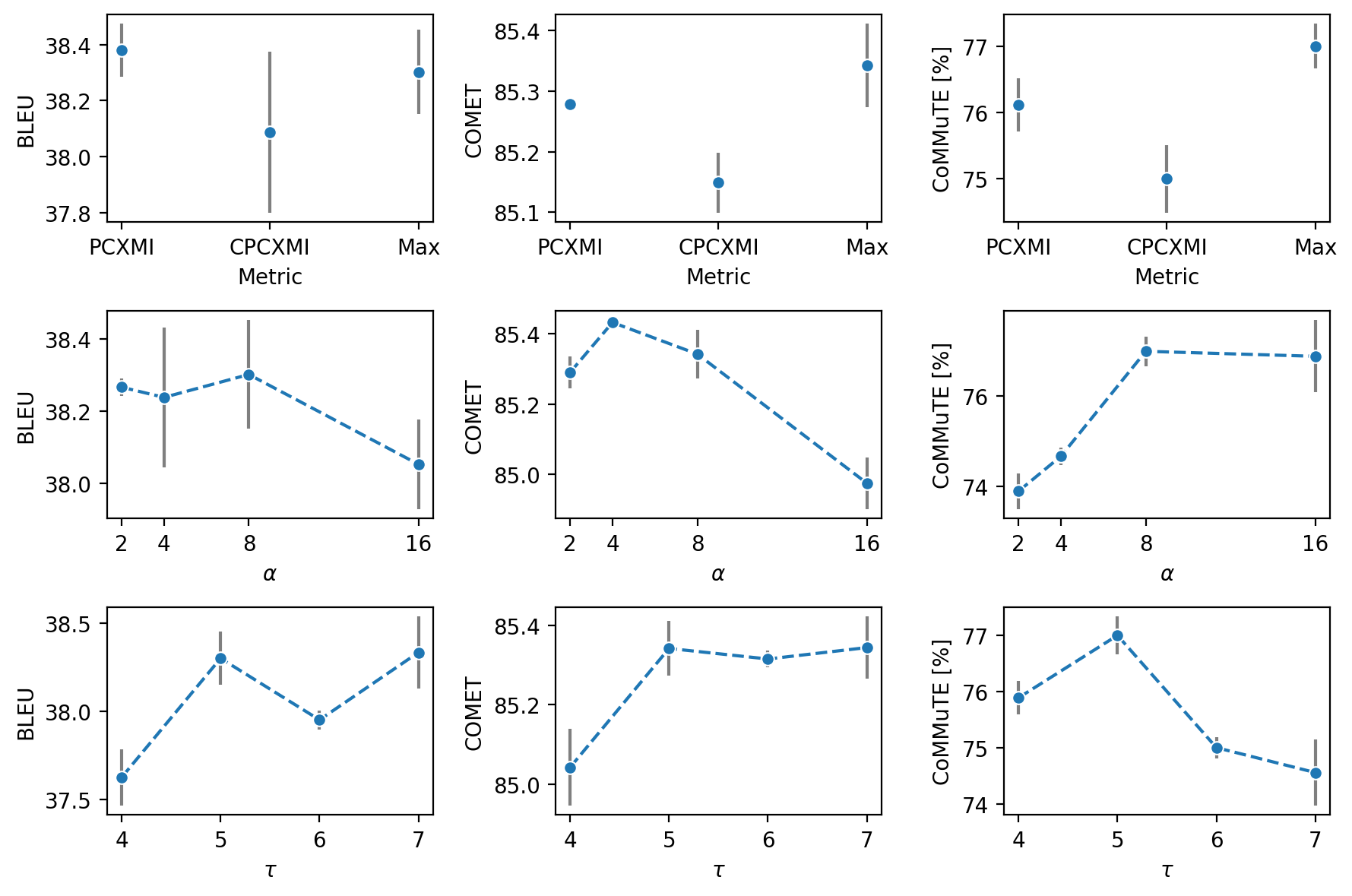}
    \caption{The results of applying the Metric-based Loss Weighting method to Qwen3-VL 4B model on the English-to-German language direction with different values of hyper-parameters. The results are in terms of the BLEU and COMET scores averaged over all used testsets, and the accuracy on CoMMuTE dataset. We averaged the results over three separate runs and error bars show the standard error.}
    \label{fig:ablation}
\end{figure*}

In Figure~\ref{fig:ablation}, we present the results of the ablation study of the Metric-based Loss Weighting method in terms of BLEU and COMET scores averaged over all testsets, and CoMMuTE accuracy. We trained Qwen3-VL 4B model using our method with different hyper-parameters: type of metric (PCXMI, CPCXMI, and their maximum), $\alpha \in \{2, 4, 8, 16\}$, and $\tau \in \{4, 5, 6, 7\}$. The results show that using the maximum of both metrics yield the higher CoMMuTE accuracy than using the metrics separately. The results for the $\alpha$ and $\tau$ hyper-parameters are stable, but we make the following observations. Increasing the $\alpha$ parameter lead to the increase in the CoMMuTE accuracy but can reduce the BLEU and COMMET scores when set too high. High $\tau$ values result in high BLEU and COMET scores, but lower CoMMuTE accuracies. This is expected as $\tau$ acts as a threshold for the token selection, and high values correspond to the lower number of tokens with increased loss.

\section{Extended Metrics Analysis}
\label{sec:metrics-analysis-extended}

\begin{figure*}[h]
\centering
\small
\begin{tabular}{ll}
\multirowcell{6}[0pt][r]{\includegraphics[height=2.6cm]{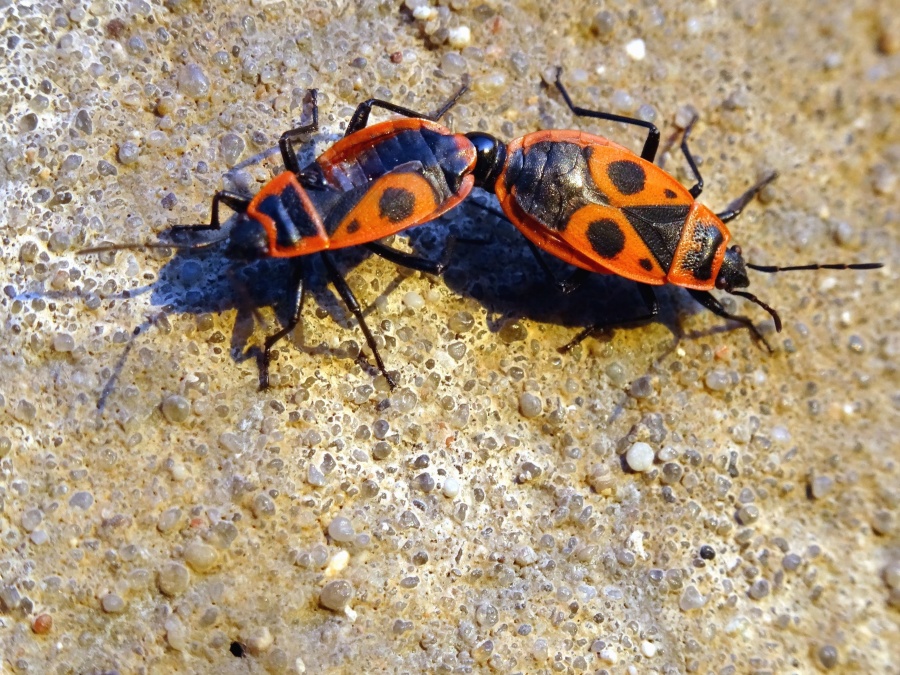}} & English: There are still a couple of \incorrect{bugs} left. \\
~ & German: Es gibt noch ein paar \incorrect{Insekten}. \\
~ & \multirowcell{3}[0pt][l]{\includegraphics[width=9.2cm]{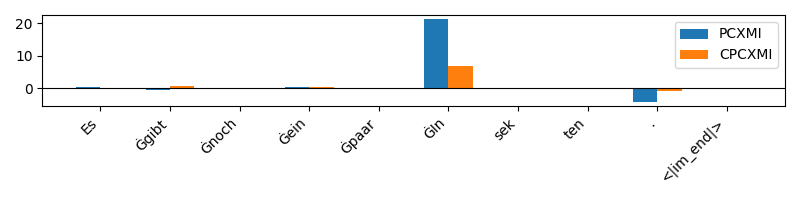}} \\
~ & ~ \\
~ & ~ \\
~ & ~ \\
~ & ~ \\
~ & ~ \\
\multirowcell{6}[0pt][r]{\includegraphics[height=2.6cm]{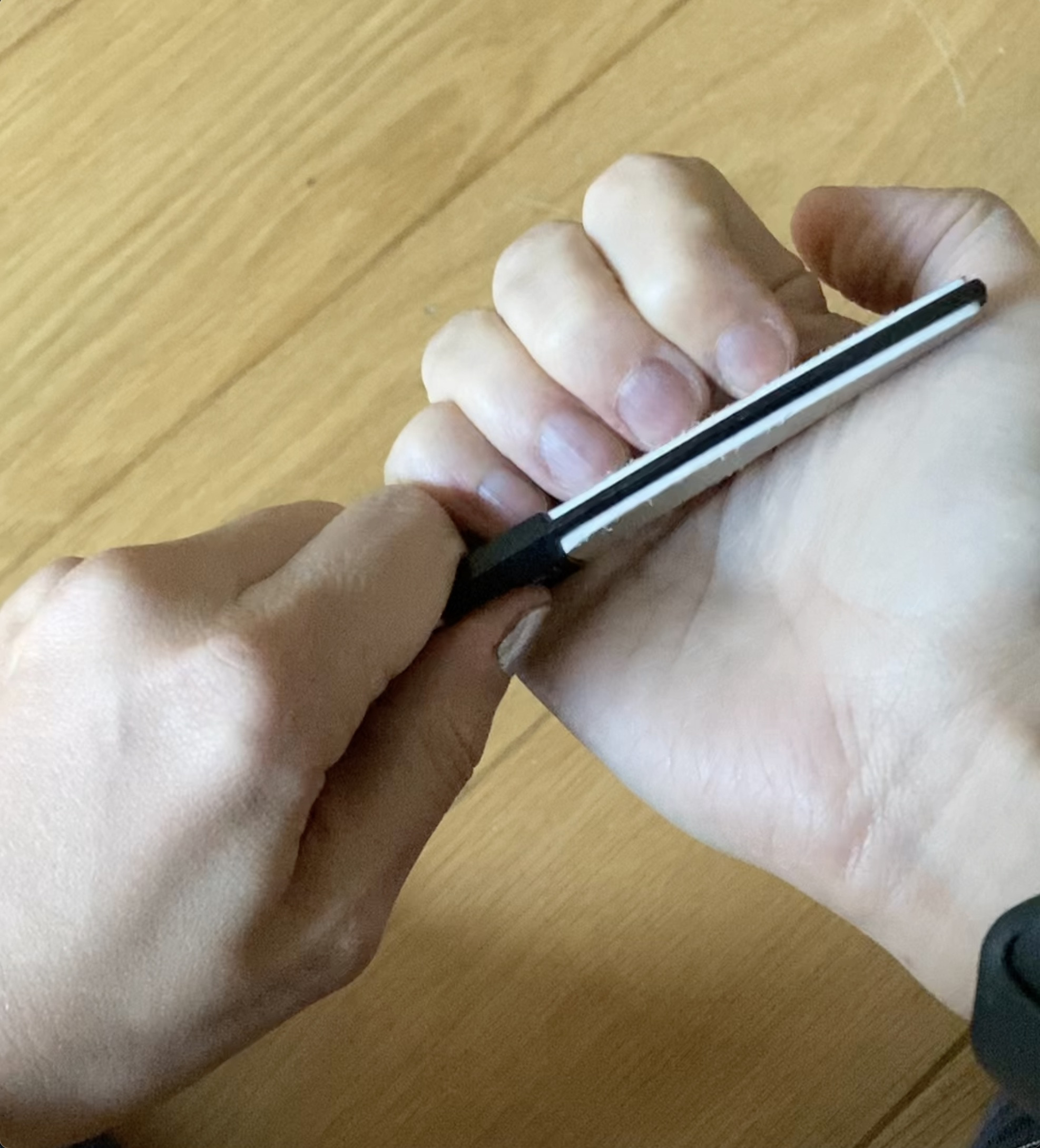}} & English: Give me the \incorrect{file} please. \\
~ & German: Geben Sie mir bitte die \incorrect{Nagelfeile}. \\
~ & \multirowcell{3}[0pt][l]{\includegraphics[width=9.2cm]{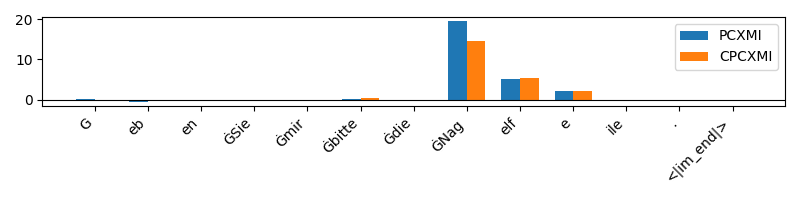}} \\
~ & ~ \\
~ & ~ \\
~ & ~ \\
~ & ~ \\
~ & ~ \\
\multirowcell{6}[0pt][r]{\includegraphics[height=2.6cm]{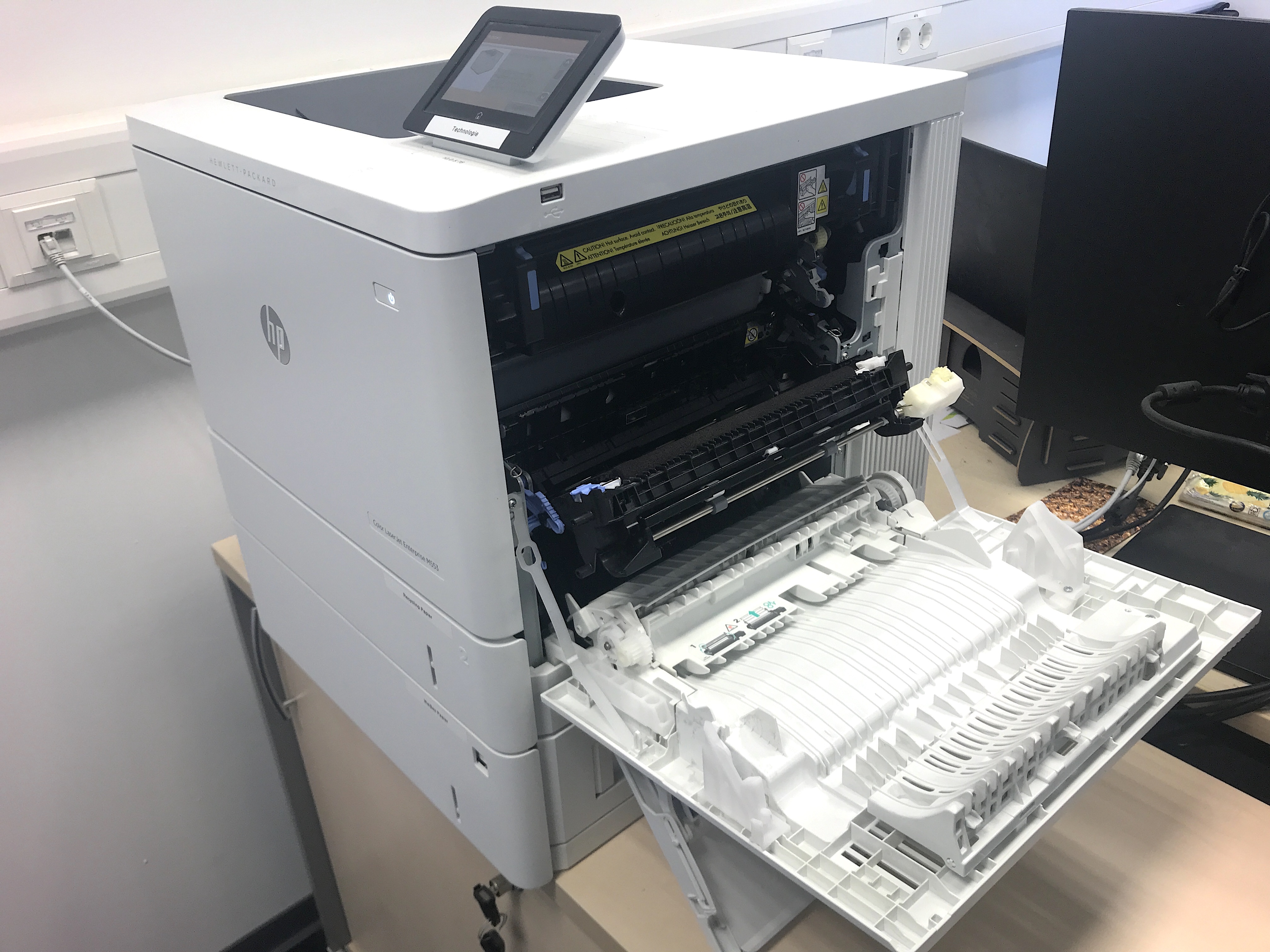}} & English: We've got another \incorrect{jam} I'm afraid. \\
~ & German: Ich fürchte, wir haben wieder einen \incorrect{Papierstau}. \\
~ & \multirowcell{3}[0pt][l]{\includegraphics[width=9.7cm]{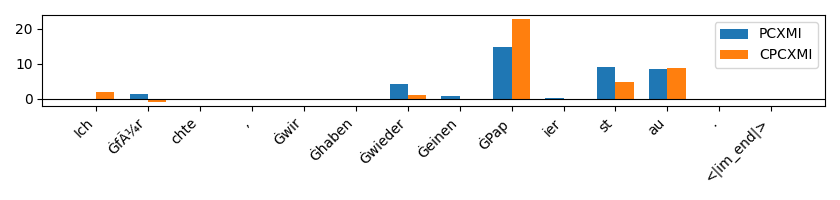}} \\
~ & ~ \\
~ & ~ \\
~ & ~ \\
~ & ~ \\
~ & ~ \\
\multirowcell{6}[0pt][r]{\includegraphics[height=2.6cm]{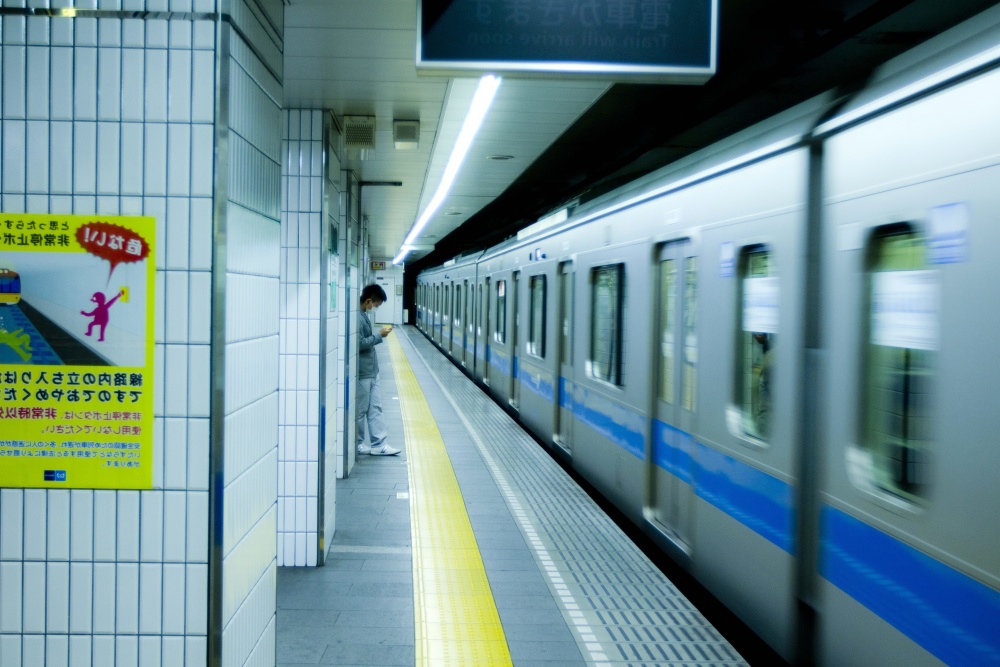}} & English: The \incorrect{tube} looks dirty! \\
~ & German: Die \incorrect{U-Bahn} sieht dreckig aus! \\
~ & \multirowcell{3}[0pt][l]{\includegraphics[width=9.2cm]{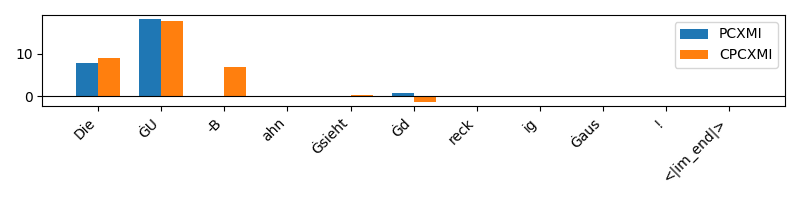}} \\
~ & ~ \\
~ & ~ \\
~ & ~ \\
~ & ~ \\
~ & ~ \\
\multirowcell{6}[0pt][c]{\includegraphics[height=2.6cm]{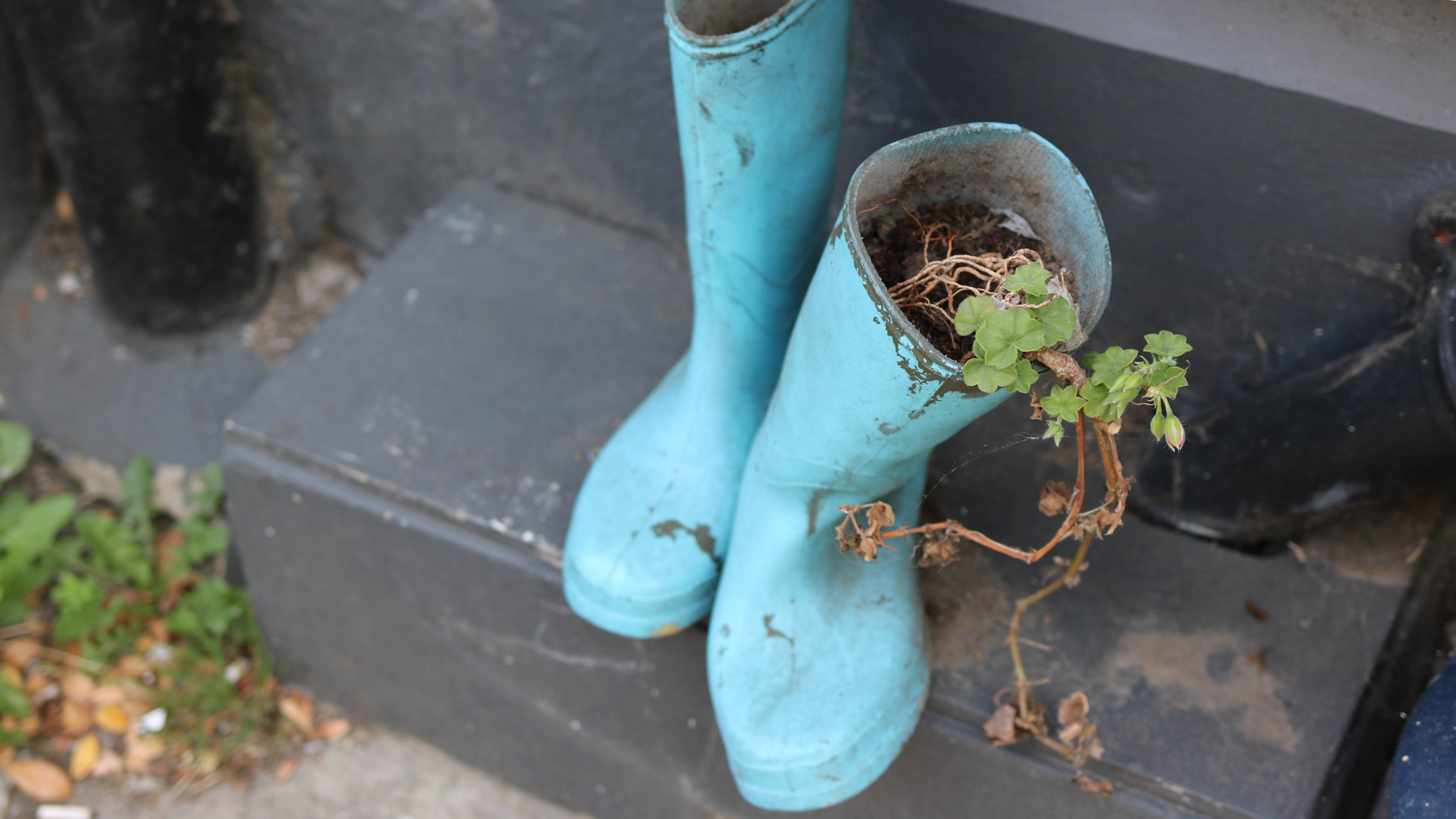}} & English: First we should fill up the \incorrect{boot}. \\
~ & German: Zuerst müssen Sie den \incorrect{Stiefel} füllen. \\
~ & \multirowcell{3}[0pt][l]{\includegraphics[width=9.7cm]{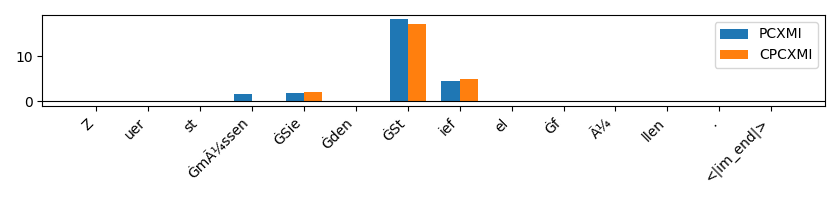}} \\
~ & ~ \\
~ & ~ \\
~ & ~ \\
~ & ~ \\
~ & ~ \\
\multirowcell{6}[0pt][r]{\includegraphics[height=2.6cm]{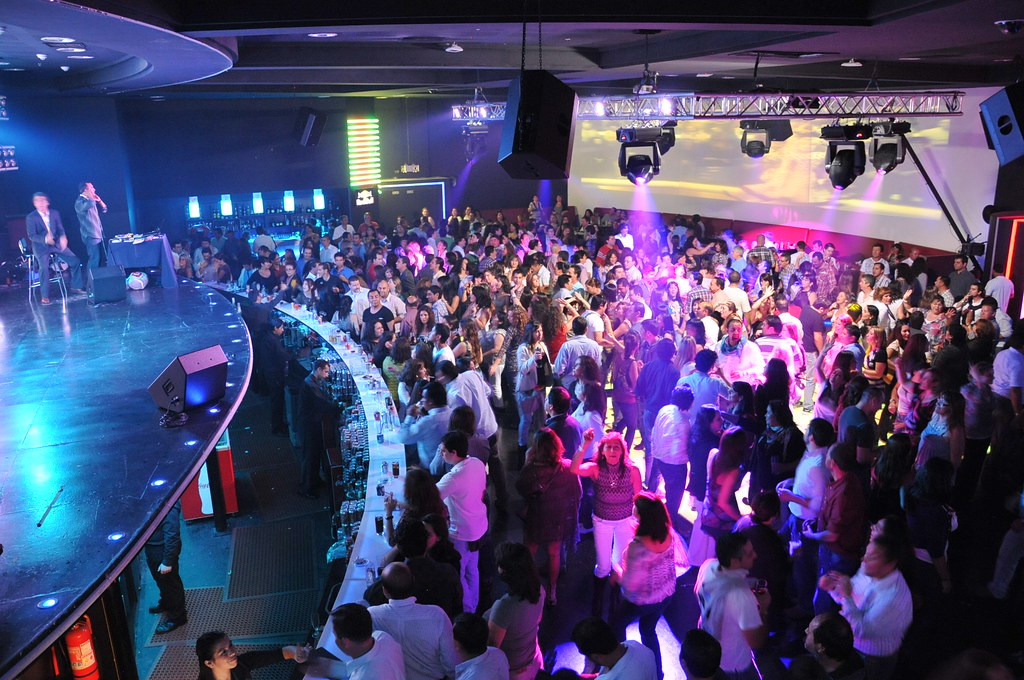}} & English: I have a \incorrect{club}. \\
~ & German: Ich habe einen \incorrect{Nachtclub}. \\
~ & \multirowcell{3}[0pt][l]{\includegraphics[width=9cm]{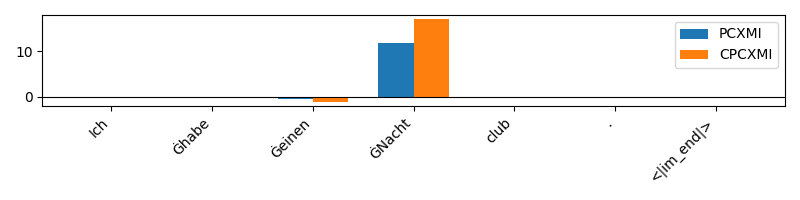}} \\
~ & ~ \\
~ & ~ \\
~ & ~ \\
~ & ~ \\
~ & ~ \\
\multirowcell{6}[0pt][r]{\includegraphics[height=2.6cm]{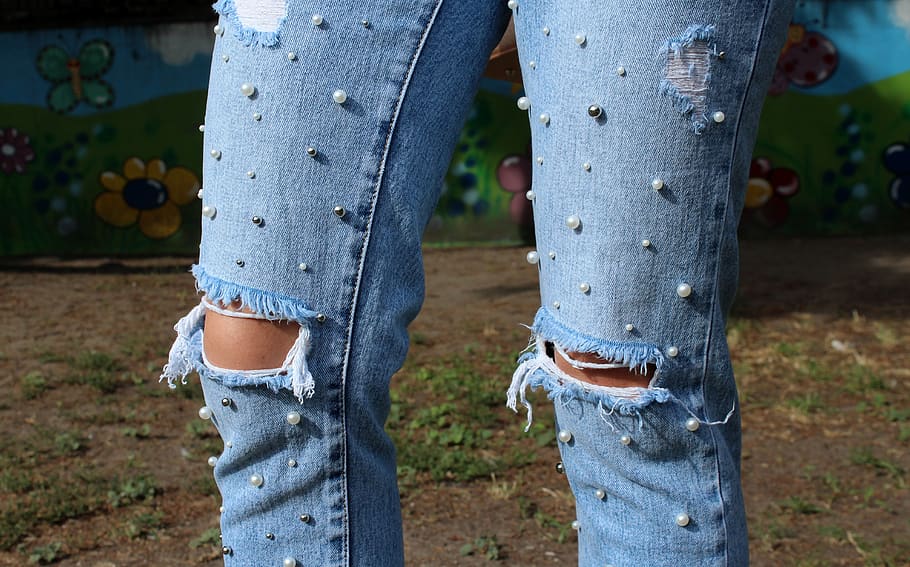}} & English: Just a single \incorrect{tear}, but it was enough. \\
~ & German: Es war nur ein einziger \incorrect{Riss}, aber er reichte aus. \\
~ & \multirowcell{3}[0pt][l]{\includegraphics[width=10cm]{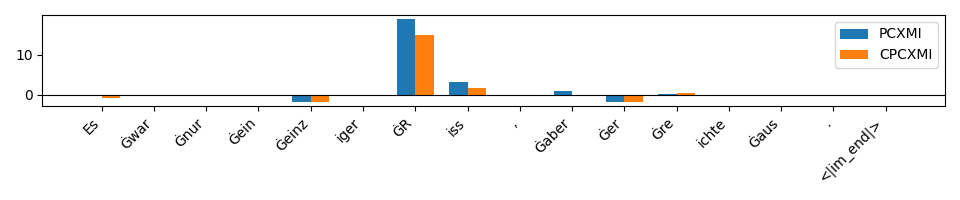}} \\
~ & ~ \\
~ & ~ \\
~ & ~ \\
~ & ~ \\
~ & ~ \\
\multirowcell{6}[0pt][r]{\includegraphics[height=2.6cm]{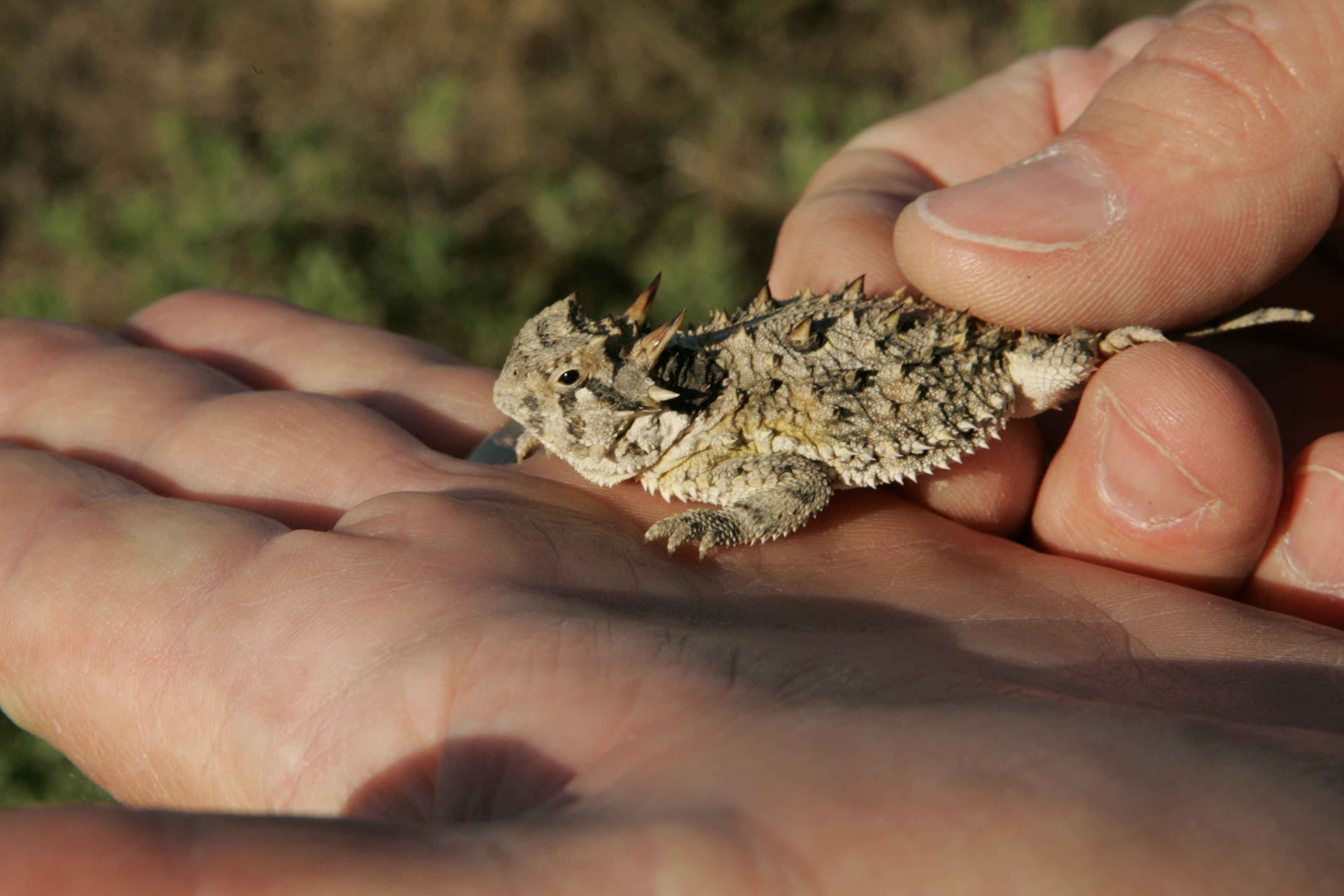}} & English: An animal sitting in a \incorrect{palm}. \\
~ & German: Ein Tier, das in einem \incorrect{Handteller} sitzt. \\
~ & \multirowcell{3}[0pt][l]{\includegraphics[width=9cm]{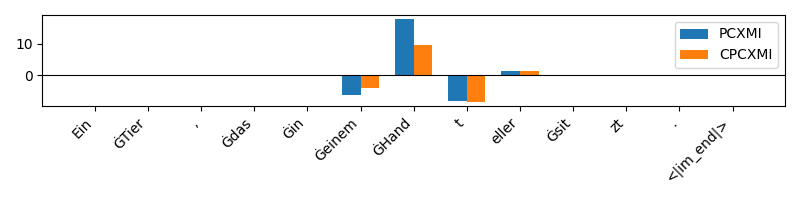}} \\
~ & ~ \\
~ & ~ \\
~ & ~ \\
~ & ~ \\
~ & ~ \\
\end{tabular}
\caption{Selected examples from the English-to-German CoMMuTE dataset with the per-token values of PCXMI and CPCXMI metrics for the Qwen3-VL 4B model. In \incorrect{red} we mark the ambiguous words the source side and the corresponding words on the target side.}
\label{fig:pcxmi-examples-expanded}
\end{figure*}

In Figure~\ref{fig:pcxmi-examples-expanded}, we present more examples of per-token values of the metrics from the CoMMuTE dataset. The metrics are capable of correctly identifying the image-dependent tokens in the target sentences.

\begin{figure*}[ht]
    \centering
    \includegraphics[width=1.0\linewidth]{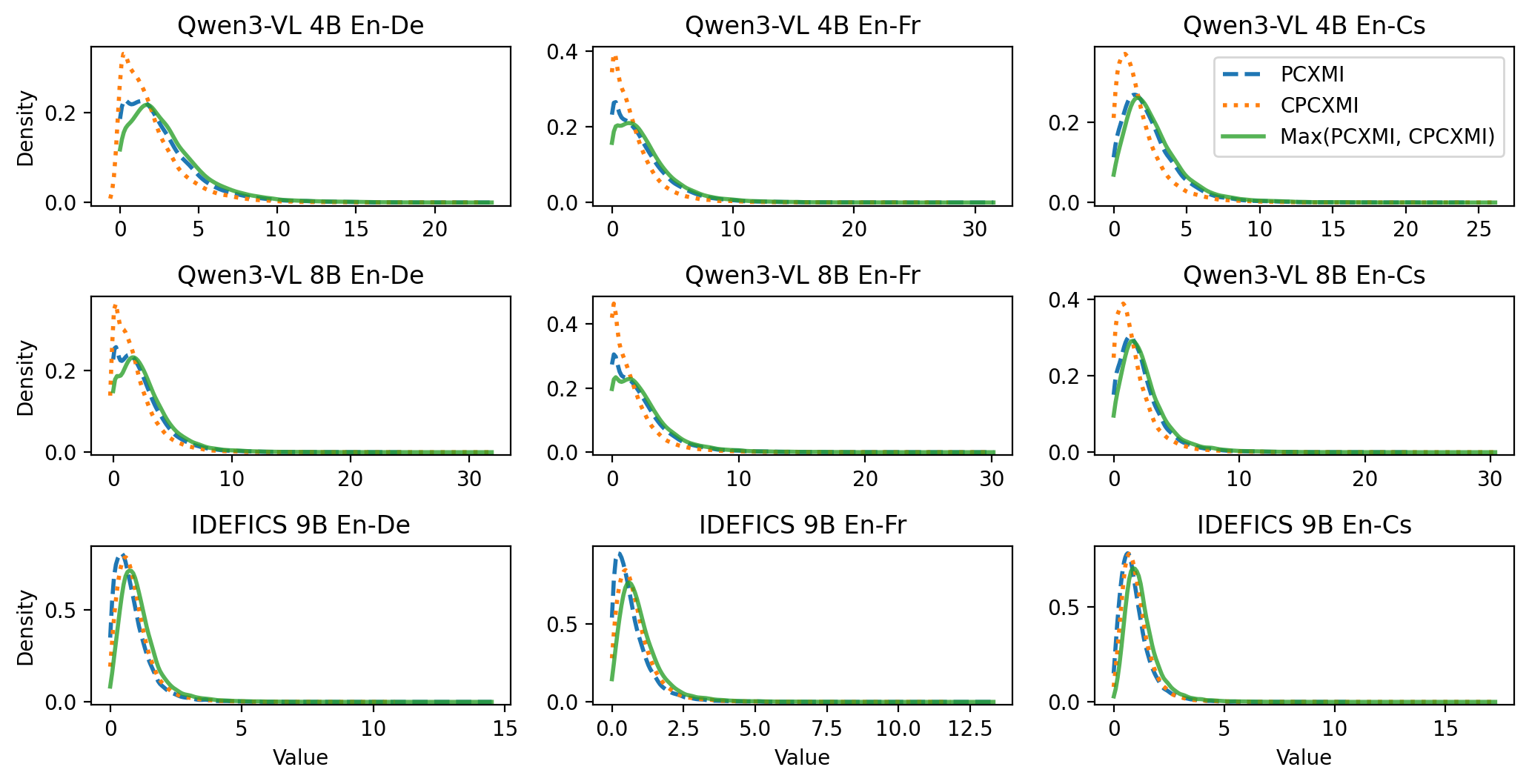}
    \caption{Kernel Density Estimate visualization of distributions of \textbf{metrics (PCXMI, CPCXMI, and the maximum of both)} in the Multi30k training dataset for each tested model and language direction. We take the maximum token-level metric for each example in the dataset.}
    \label{fig:pcxmi-distribution}
\end{figure*}

Figure~\ref{fig:pcxmi-distribution} shows the distribution of values of the metrics (PCXMI, CPCXMI and the maximum of both) on the Multi30k training dataset. We take the maximum value of a metric of the target tokens for each example in the dataset. For both Qwen3-VL models and all language directions, the CPCXMI is lower than PCXMI, showing higher density for values close to zero. This is not the case for IDEFICS 9B, where the distributions are very similar for all language directions. 

\begin{figure*}[ht]
    \centering
    \includegraphics[width=1.0\linewidth]{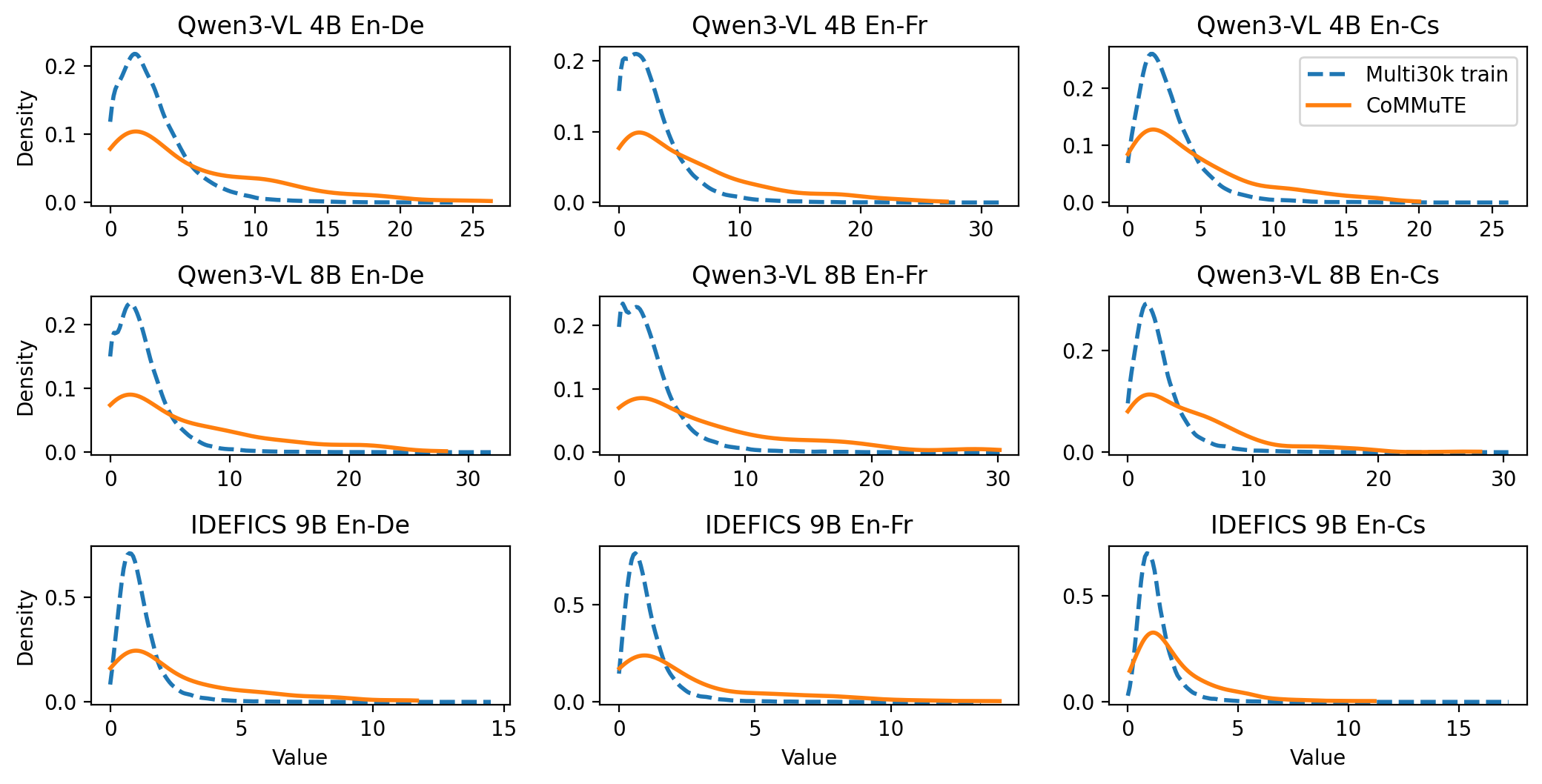}
    \caption{Kernel Density Estimate visualization of distributions of Maximum of PCXMI and CPCXMI in the \textbf{Multi30k training and CoMMuTE datasets} for each tested model and language direction. We take the maximum token-level metric for each example in the dataset.}
    \label{fig:pcxmi-distribution-comparison}
\end{figure*}

The comparison between the distributions of the maximum of PCXMI and CPCXMI on the training split of Multi30k and the CoMMuTE dataset is shown in Figure~\ref{fig:pcxmi-distribution-comparison}. The metric for CoMMuTE is not as concentrated close to zero as for Multi30k, which is consistent with the presence of ambiguous words in every example of the CoMMuTE dataset.

\section{Extended Qualitative Analysis}
\label{sec:extended-qualitative-analysis}

\begin{figure*}[h]
\centering
\small
\begin{tabular}{ll}
\multirowcell{6}[0pt][r]{\includegraphics[height=2.1cm]{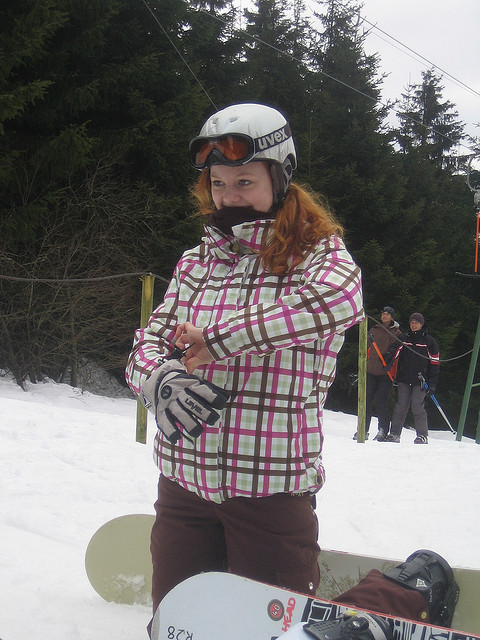}} & English: A young woman on a snowboard putting on gloves. \\
~ & French: Une jeune femme sur un snowboard mettant des gants. \\
~ & Fine-tuning: Une jeune femme sur un snowboard mettant des gants. \\
~ & Weighting: Une jeune femme \incorrect{en} snowboard \incorrect{en train de} mettre des gants. \\
~ & BLEU: 100.00/22.03 (-77.97) | COMET: 96.36/86.10 (-10.26) \\
~ & Weighting used present continuous. \\
~ & ~ \\
\multirowcell{6}[0pt][r]{\includegraphics[height=2.1cm]{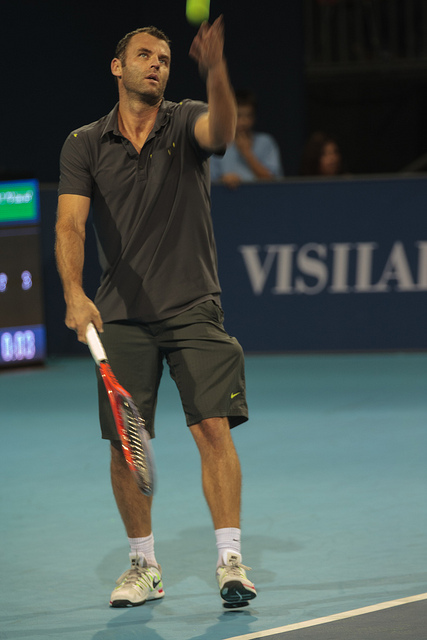}} & English: A person that is playing in a tennis game. \\
~ & French: Une personne qui joue un match de tennis. \\
~ & Fine-tuning: Une personne qui joue un match de tennis. \\
~ & Weighting: Une personne qui joue au tennis. \\
~ & BLEU: 100.00/36.74 (-63.26) | COMET: 97.80/81.47 (-16.33) \\
~ & Weighting did not include the tennis match. \\
~ & ~ \\
\multirowcell{6}[0pt][r]{\includegraphics[height=2.1cm]{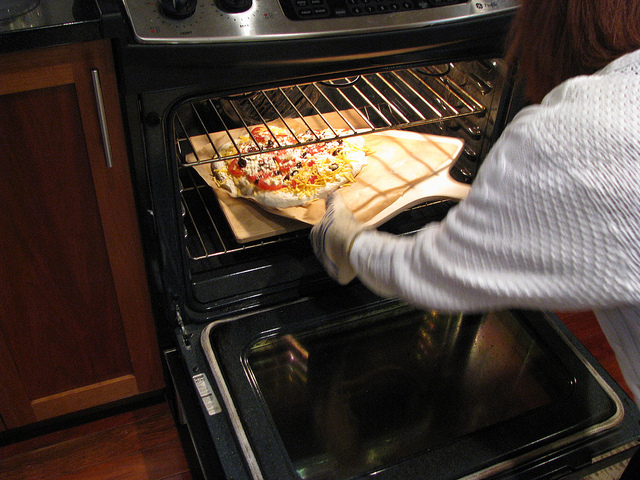}} & English: A man pulling a pizza out of an oven. \\
~ & French: Un homme sortant une pizza d'un four. \\
~ & Fine-tuning: Un homme sortant une pizza d'un four. \\
~ & Weighting: Un homme \incorrect{en train de} sortir une pizza d'un four. \\
~ & BLEU: 100.00/40.35 (-59.65) | COMET: 97.05/92.57 (-4.48) \\
~ & Weighting used present continuous. \\
~ & ~ \\
\multirowcell{6}[0pt][r]{\includegraphics[height=2.1cm]{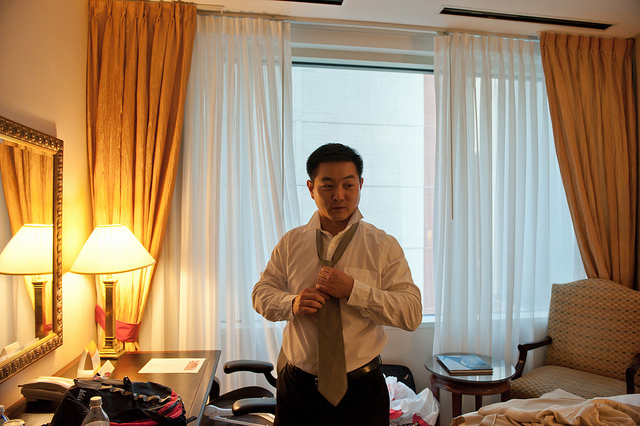}} & English: There is a man putting on a tie. \\
~ & French: Il y a un homme qui met une cravate. \\
~ & Fine-tuning: Il y a un homme qui met une cravate. \\
~ & Weighting: Il y a un homme \incorrect{en train de} mettre une cravate. \\
~ & BLEU: 100.00/42.40 (-57.60) | COMET: 95.79/79.28 (-16.51) \\
~ & Weighting used present continuous. \\
~ & ~ \\
\multirowcell{6}[0pt][r]{\includegraphics[height=2.1cm]{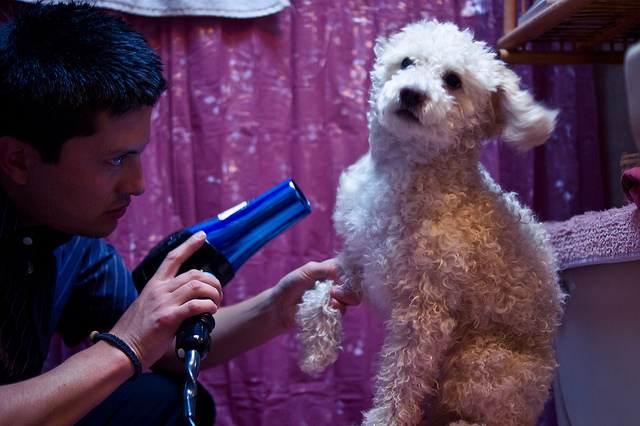}} & English: A dog being blown dry by a man. \\
~ & French: Un chien séché par un homme. \\
~ & Fine-tuning: Un chien séché par un homme. \\
~ & Weighting: Un chien \incorrect{en train d}'être séché par un homme. \\
~ & BLEU: 100.00/45.18 (-54.82) | COMET: 95.76/75.62 (-20.14) \\
~ & Weighting used present continuous. \\
~ & ~ \\
\multirowcell{6}[0pt][r]{\includegraphics[height=2.1cm]{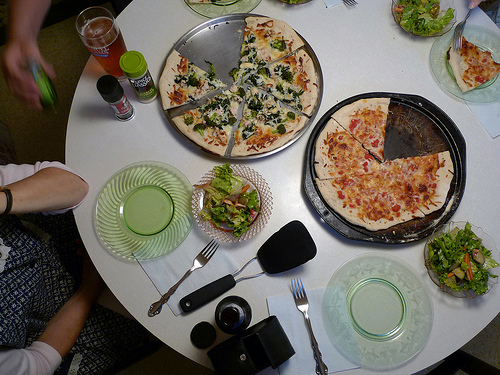}} & English: A table set for a dinner of pizza. \\
~ & French: Une table dressée pour un dîner de pizza. \\
~ & Fine-tuning: Une table dressée pour un dîner de pizza. \\
~ & Weighting: Une table dressée pour un dîner \incorrect{avec} des pizzas. \\
~ & BLEU: 100.00/53.73 (-46.27) | COMET: 97.67/92.51 (-5.16) \\
~ & Weighting translated as "dinner with pizzas". \\
~ & ~ \\
\multirowcell{6}[0pt][r]{\includegraphics[height=2.1cm]{images/examples/COCO_train2014_000000141557.jpg}} & English: A man is reading a book while another is holding a dog. \\
~ & French: Un homme est en train de lire un livre pendant qu'un autre tient un chien. \\
~ & Fine-tuning: Un homme est en train de lire un livre tandis qu'un autre tient un chien. \\
~ & Weighting: Un homme lit un livre tandis qu'un autre tient un chien. \\
~ & BLEU: 81.54/36.95 (-44.59) | COMET: 92.98/86.83 (-6.15) \\
~ & Weighting used present simple. \\
~ & ~ \\
\multirowcell{6}[0pt][r]{\includegraphics[height=2.1cm]{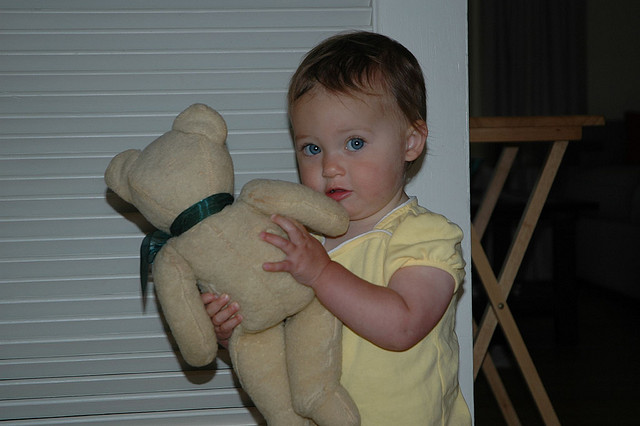}} & English: Baby in yellow holds on to her teddy bear. \\
~ & French: Le bébé en jaune tient son ours en peluche. \\
~ & Fine-tuning: Un bébé en jaune tient son ours en peluche. \\
~ & Weighting: Un bébé en jaune \incorrect{se} tient \incorrect{à} son ours en peluche. \\
~ & BLEU: 88.01/43.67 (-44.34) | COMET: 92.68/85.95 (-6.73) \\
~ & Weighting changed the meaning to "supporting oneself". \\
~ & ~ \\
\multirowcell{6}[0pt][r]{\includegraphics[height=2.1cm]{images/examples/COCO_val2014_000000527029.jpg}} & English: A man covering his eyes next to cases of bananas \\
~ & French: Un homme couvrant ses yeux à côté des boîtiers de bananes \\
~ & Fine-tuning: Un homme couvrant ses yeux à côté de caisses de bananes \\
~ & Weighting: Un homme \incorrect{se} couvrant \incorrect{les} yeux à côté de caisses de bananes \\
~ & BLEU: 63.16/19.16 (-44.00) | COMET: 89.27/85.48 (-3.79) \\
~ & Weighting used the reflexive form. \\
~ & ~ \\
\end{tabular}\caption{Selected examples with the highest difference in BLEU between Fine-tuning and Metric-based Loss Weighting (Fine-tuning scoring higher) in the MSCOCO testset on the English-to-French language direction of Qwen3-VL 8B. We also report BLEU and COMET scores for the translations by the methods (and their differences), and our description of errors.}
\label{tab:qualitative-top-difference}
\end{figure*}

\begin{figure*}[h]
\centering
\small
\begin{tabular}{ll}
\multirowcell{6}[0pt][r]{\includegraphics[height=2.1cm]{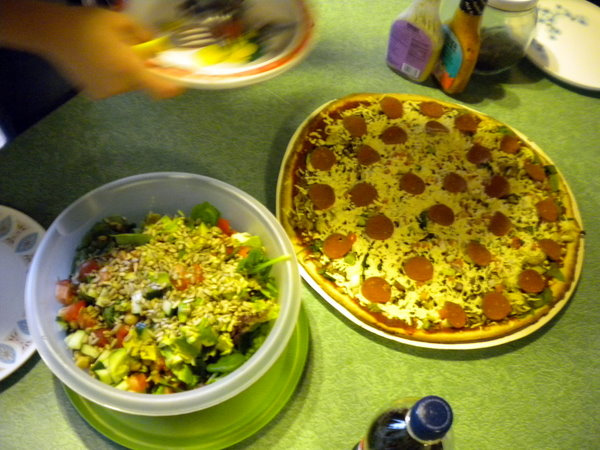}} & English: A pizza sitting next to a bowl of salad. \\
~ & French: Une pizza posée à côté d'un bol de salade. \\
~ & Fine-tuning: Une pizza placée à côté d'une salade. \\
~ & Weighting: Une pizza posée à côté d'un bol de salade. \\
~ & BLEU: 14.90/100.00 (85.10) | COMET: 88.46/97.40 (8.94) \\
~ & Fine-tuning ommited the bowl (of salad) \\
~ & ~ \\
\multirowcell{6}[0pt][r]{\includegraphics[height=2.1cm]{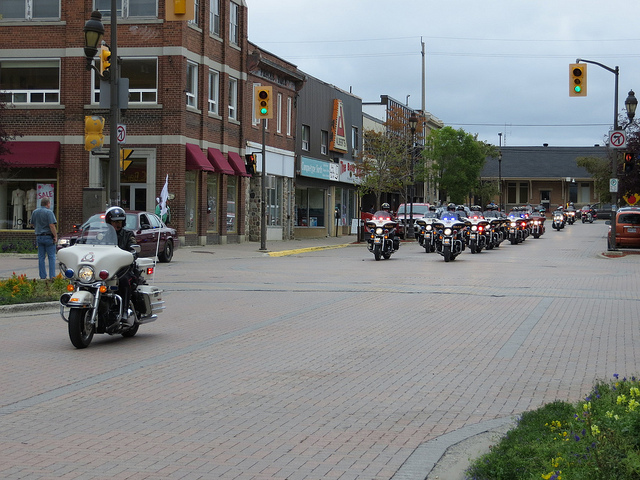}} & English: A cop leads the way for other people on motorcycles. \\
~ & French: Un policier ouvre la voie à d'autres personnes sur des motos. \\
~ & Fine-tuning: Un policier \incorrect{dirige} la \incorrect{route} pour d'autres personnes en motos. \\
~ & Weighting: Un policier ouvre la voie pour d'autres personnes sur des motos. \\
~ & BLEU: 12.32/73.49 (61.17) | COMET: 85.39/93.69 (8.30) \\
~ & Fine-tuning used the incorrect verb "to direct". \\
~ & ~ \\
\multirowcell{6}[0pt][r]{\includegraphics[height=2.1cm]{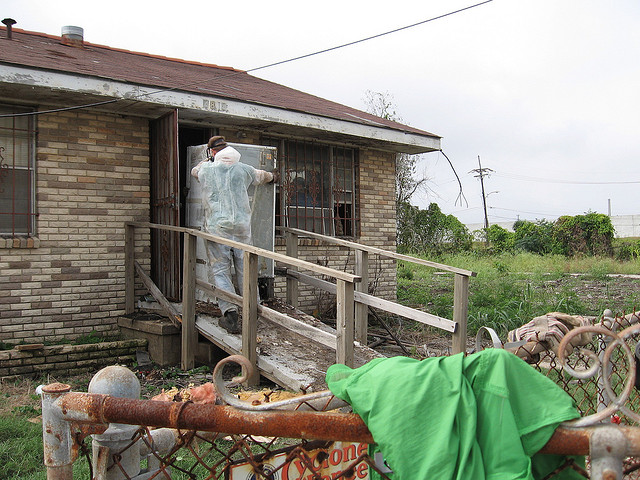}} & English: a man moving a refrigerator  in to a house \\
~ & French: un homme transportant un réfrigérateur dans d'une maison \\
~ & Fine-tuning: Un homme \incorrect{en train de} mettre un réfrigérateur dans une maison \\
~ & Weighting: un homme transportant un réfrigérateur dans une maison \\
~ & BLEU: 15.85/70.71 (54.86) | COMET: 78.21/92.69 (14.48) \\
~ & Fine-tuning used present continuous and a synonym. \\
~ & ~ \\
\multirowcell{6}[0pt][r]{\includegraphics[height=2.1cm]{images/examples/COCO_train2014_000000303592.jpg}} & English: A black cat climbing into a bath tub. \\
~ & French: Un chat noir grimpant dans une baignoire. \\
~ & Fine-tuning: Un chat noir \incorrect{montant} dans une baignoire. \\
~ & Weighting: Un chat noir grimpant dans une baignoire. \\
~ & BLEU: 50.00/100.00 (50.00) | COMET: 86.99/96.85 (9.86) \\
~ & Fine-tuning used a synonym. \\
~ & ~ \\
\multirowcell{6}[0pt][r]{\includegraphics[height=2.1cm]{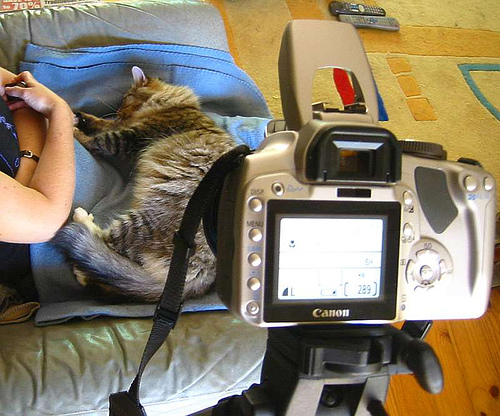}} & English: A camera set up in front of a sleeping cat. \\
~ & French: Un appareil photo installé devant un chat qui dort. \\
~ & Fine-tuning: \incorrect{Une caméra} installée devant un chat endormi. \\
~ & Weighting: Un appareil photo installé devant un chat endormi. \\
~ & BLEU: 17.20/67.17 (49.97) | COMET: 72.14/73.10 (0.96) \\
~ & Fine-tuning used incorrect "video camera". \\
~ & ~ \\
\multirowcell{6}[0pt][r]{\includegraphics[height=2.1cm]{images/examples/COCO_val2014_000000011696.jpg}} & English: A dog climbing on top of a person sleeping on a couch. \\
~ & French: Un chien qui grimpe sur une personne qui dort sur un canapé. \\
~ & Fine-tuning: Un chien \incorrect{escaladant} un \incorrect{individu} dormant sur un canapé. \\
~ & Weighting: Un chien grimpant sur une personne qui dort sur un canapé. \\
~ & BLEU: 23.14/70.77 (47.62) | COMET: 71.39/90.27 (18.88) \\
~ & Fine-tuning used incorrect in the context "climbing (e.g., on a mountain)" and a synonym. \\
~ & ~ \\
\multirowcell{6}[0pt][r]{\includegraphics[height=2.1cm]{images/examples/COCO_val2014_000000204329.jpg}} & English: A blue, yellow and green surfboard sticking out of a sandy beach. \\
~ & French: Une planche de surf bleue, jaune et verte plantée dans le sable d'une plage. \\
~ & Fine-tuning: Une planche de surf bleue, jaune et verte \incorrect{dépassant} d'une plage de sable. \\
~ & Weighting: Une planche de surf bleue, jaune et verte plantée dans le sable d'une plage. \\
~ & BLEU: 58.22/100.00 (41.78) | COMET: 68.71/95.68 (26.97) \\
~ & Fine-tuning used incorrect in the context "stick out". \\
~ & ~ \\
\multirowcell{6}[0pt][r]{\includegraphics[height=2.1cm]{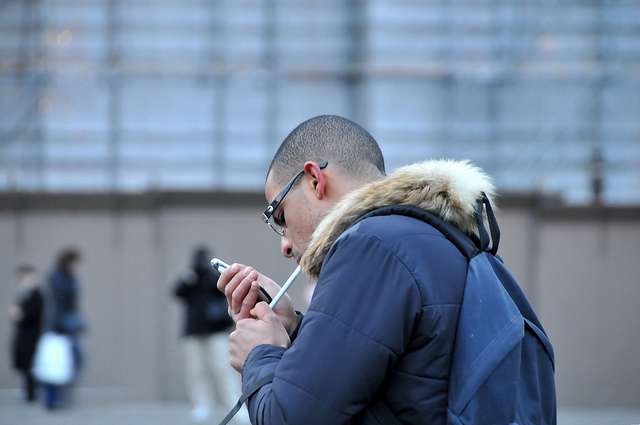}} & English: A man in a blue jacket lighting a cigarette. \\
~ & French: Un homme en veste bleue allumant une cigarette. \\
~ & Fine-tuning: Un homme en veste bleue \incorrect{allume} une cigarette. \\
~ & Weighting: Un homme en veste bleue allumant une cigarette. \\
~ & BLEU: 59.69/100.00 (40.31) | COMET: 92.68/97.31 (4.63) \\
~ & Fine-tuning used present simple. \\
~ & ~ \\
\multirowcell{6}[0pt][r]{\includegraphics[height=2.1cm]{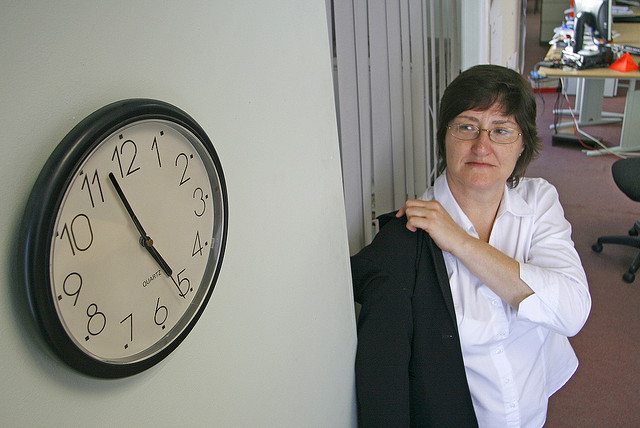}} & English: A woman putting on a jacket while looking at a clock \\
~ & French: Une femme mettant une veste en regardant une horloge \\
~ & Fine-tuning: Une femme \incorrect{en train de} mettre une veste tout en regardant une horloge \\
~ & Weighting: Une femme mettant une veste tout en regardant une horloge \\
~ & BLEU: 26.13/65.80 (39.67) | COMET: 87.06/94.40 (7.34) \\
~ & Fine-tuning used present continuous. \\
~ & ~ \\
\end{tabular}\caption{Selected examples with the highest difference in BLEU between Metric-based Loss Weighting and Fine-tuning (Weighting scoring higher) in the MSCOCO testset on the English-to-French language direction of Qwen3-VL 8B. We also report BLEU and COMET scores for the translations by the methods (and their differences), and our description of errors.}
\label{tab:qualitative-bottom-difference}
\end{figure*}

To examine the source of the largest measured drop in BLEU between Fine-tuning and Metric-based Loss Weighting in our experiments, we analyzed the translations of Qwen3-VL 8B on \textit{MSCOCO} English-to-French testset of the runs with seed 1, which also shows the highest BLEU difference. We looked at the 20 examples with the highest difference of example-level BLEU scores between the two methods (20 in each direction). The selected instances for Fine-tuning scoring higher can be seen in Figure~\ref{tab:qualitative-top-difference}, and in Figure~\ref{tab:qualitative-bottom-difference} for Weighting scoring higher. We categorize the causes of the drop in BLEU into three categories: (1) change of the tense (mostly present continuous instead of present simple), (2) synonyms or equivalent expressions, (3) true errors including omissions and incorrect meaning. Out of the 20 instances with highest drop of BLEU for Weighting 30\% were due to the use of a different form, 45\% were equivalent, and 25\% were erroneous. From the 20 instances where Fine-tuning showed the largest drop in BLEU 10\% were due to a different form, 50\% were synonyms, and 40\% contained errors. While this analysis only looked at the selected examples, it shows that the largest drop in example-level BLEU scores mostly corresponds to acceptable translations that differ from the gold target translation.

\section{Extended Results}
\label{sec:extended-results}

In this section, we present the full results for the combined method (Metric-based Loss Weighting and Word Dropout). Table~\ref{tab:combined-full} shows the results for all testsets on the English-to-German and English-to-French language directions for the Qwen3-VL 4B model.

\begin{table*}[ht]
\centering
\tiny
\begin{tabular}{lrrrrrr|rrr} 
\hline
\textbf{Model} & \multicolumn{2}{c}{\textbf{test 2016}} & \multicolumn{2}{c}{\textbf{test 2017}} & \multicolumn{2}{c|}{\textbf{MSCOCO}} & \multicolumn{2}{c}{\textbf{Average}} & \textbf{CoMMuTE} \\ 
~ & \textbf{BLEU} & \textbf{COMET} & \textbf{BLEU} & \textbf{COMET} & \textbf{BLEU} & \textbf{COMET} & \textbf{BLEU} & \textbf{COMET} & \textbf{Accuracy} \\ 
\hline
\multicolumn{10}{c}{\textbf{English-to-German}} \\
\hline
Qwen3-VL 4B & 38.51 & 85.56 & 32.95 & 84.90 & 29.94 & 81.24 & 33.80 & 83.90 & 74.00\% \\ 
~~~Fine-tuning & \textbf{42.61}$\pm$0.12 & 86.66$\pm$0.06 & \textbf{38.08}$\pm$0.38 & 85.98$\pm$0.06 & \underline{35.22}$\pm$0.45 & 82.70$\pm$0.13 & \textbf{38.64}$\pm$0.18 & 85.11$\pm$0.05 & 71.67$\pm$0.19\% \\ 
~~~Word Dropout & \underline{42.20}$\pm$0.19 & 86.62$\pm$0.06 & 37.17$\pm$0.19 & 85.80$\pm$0.01 & 35.00$\pm$0.22 & 82.69$\pm$0.16 & 38.13$\pm$0.18 & 85.03$\pm$0.04 & 76.89$\pm$0.44\% \\ 
~~~Metric Weighting & 41.91$\pm$0.33 & \underline{86.80}$\pm$0.04 & \underline{37.29}$\pm$0.14 & \textbf{86.14}$\pm$0.04 & \textbf{35.71}$\pm$0.19 & \textbf{83.09}$\pm$0.13 & \underline{38.30}$\pm$0.15 & \textbf{85.34}$\pm$0.07 & \underline{77.00}$\pm$0.33\% \\ 
~~~Weighting + Dropout & 41.56$\pm$0.49 & \textbf{86.87}$\pm$0.07 & 36.85$\pm$0.36 & \underline{86.04}$\pm$0.07 & 34.99$\pm$0.06 & \underline{82.80}$\pm$0.19 & 37.80$\pm$0.11 & \underline{85.23}$\pm$0.09 & \textbf{78.78}$\pm$0.11\% \\
\hline
\multicolumn{10}{c}{\textbf{English-to-French}} \\
\hline
Qwen3-VL 4B & 43.77 & 86.04 & 43.47 & 86.31 & 43.24 & 83.69 & 43.49 & 85.35 & 75.00\% \\ 
~~~Fine-tuning & \underline{67.03}$\pm$0.36 & \textbf{89.81}$\pm$0.05 & \textbf{61.62}$\pm$0.13 & \textbf{89.25}$\pm$0.10 & \underline{50.90}$\pm$0.43 & \textbf{85.37}$\pm$0.12 & \textbf{59.85}$\pm$0.08 & \textbf{88.14}$\pm$0.03 & 81.17$\pm$0.37\% \\ 
~~~Word Dropout & 66.46$\pm$0.31 & 89.58$\pm$0.09 & 60.98$\pm$0.05 & 89.10$\pm$0.02 & \textbf{51.01}$\pm$0.44 & 85.15$\pm$0.04 & 59.48$\pm$0.27 & 87.94$\pm$0.04 & 84.42$\pm$0.68\% \\ 
~~~Metric Weighting & \textbf{67.22}$\pm$0.28 & \underline{89.69}$\pm$0.03 & \underline{61.04}$\pm$0.23 & \underline{89.15}$\pm$0.04 & 50.50$\pm$0.37 & 85.07$\pm$0.03 & \underline{59.59}$\pm$0.22 & \underline{87.97}$\pm$0.03 & \textbf{84.74}$\pm$0.50\% \\ 
~~~Weighting + Dropout & 65.74$\pm$0.39 & 89.49$\pm$0.10 & 59.69$\pm$0.07 & 89.12$\pm$0.07 & 50.71$\pm$0.32 & \underline{85.19}$\pm$0.19 & 58.71$\pm$0.22 & 87.93$\pm$0.09 & \underline{84.63}$\pm$0.39\% \\
\hline
\end{tabular}
\caption{Results for combined Metric-based Loss Weighting and Word Dropout for all testsets in English-to-German and English-to-French. The results are averaged over three separate runs.}
\label{tab:combined-full}
\end{table*}

\end{document}